%% file: iclr2026_conference.tex
\documentclass{article}
\usepackage{enumitem}
\usepackage{iclr2026_conference,times}

\input{math_commands.tex}

\usepackage{makecell} %
\usepackage{booktabs}
\usepackage{url}
\usepackage{graphicx}
\usepackage{amsfonts}
\usepackage{tabularx,booktabs,multirow}
\usepackage[pagebackref=true,hidelinks]{hyperref}
\usepackage{url}
\usepackage{graphicx}
\usepackage{amsfonts}
\usepackage{tabularx,booktabs,multirow}
\usepackage{float}
\usepackage{overpic}
\usepackage{xcolor}
\usepackage{wrapfig}
\usepackage{nicefrac}
\usepackage{microtype}
\usepackage{xcolor}
\usepackage{enumitem}
\usepackage{amsmath,amssymb}
\usepackage{algorithm}
\usepackage[noend]{algpseudocode}
\usepackage{float}  
\usepackage{caption}
\usepackage{titlesec}
\usepackage{listings}
\usepackage[T1]{fontenc}
\algrenewcommand{\algorithmiccomment}[1]{\hfill\(\triangleright\)~#1}

\renewcommand{\paragraph}[1]{\noindent\textbf{#1}.}
\titlespacing\section{0pt}{3pt plus 1pt minus 1pt}{2pt plus 1pt minus 1pt}
\titlespacing\subsection{0pt}{2pt plus 1pt minus 1pt}{1pt plus 1pt minus 1pt}
\graphicspath{{Figures/}}

\title{\NAME : Orchestrating Small Language Models for Reasoning}

\author{
Chenyu Wang\textsuperscript{1*\dag} \quad
Zishen Wan\textsuperscript{1,2*\dag} \quad
Hao Kang\textsuperscript{2} \quad
Emma Chen\textsuperscript{1} \\
\textbf{Zhiqiang Xie}\textsuperscript{3} \quad
\textbf{Tushar Krishna}\textsuperscript{2} \quad
\textbf{Vijay Janapa Reddi}\textsuperscript{1} \quad
\textbf{Yilun Du}\textsuperscript{1} \\[0.3em]
\textsuperscript{1}Harvard University \quad
\textsuperscript{2}Georgia Institute of Technology \quad
\textsuperscript{3}Stanford University
}

\newcommand{\homehl}[1]{#1}
\newcommand{\NAME}{\textsc{SLM-MUX}} 
\iclrfinalcopy 

\begin{document}

\maketitle
\lhead{}
\renewcommand{\headrulewidth}{0pt}

\renewcommand{\thefootnote}{\fnsymbol{footnote}}
\footnotetext[1]{Equal contribution.}
\footnotetext[2]{Correspondence to \texttt{chenyu\_wang@seas.harvard.edu}, \texttt{zishenwan@seas.harvard.edu}.}
\renewcommand{\thefootnote}{\arabic{footnote}}
\setcounter{footnote}{0}

\begin{abstract}
 With the rapid development of language models, the number of small language models (SLMs) has grown significantly. Although they do not achieve state-of-the-art accuracy, they are more efficient and often excel at specific tasks. This raises a natural question: can multiple SLMs be orchestrated into a system where each contributes effectively, achieving higher accuracy than any individual model? Existing orchestration methods have primarily targeted frontier models (e.g., GPT-4) and perform suboptimally when applied to SLMs. To address this gap, we propose a three-stage approach for orchestrating SLMs. First, we introduce \NAME{}, a multi-model architecture that effectively coordinates multiple SLMs. Building on this, we develop two optimization strategies: (i) a model selection search that identifies the most complementary SLMs from a given pool, and (ii) test-time scaling tailored to \NAME{}. Our approach delivers strong results: Compared to existing orchestration methods, our approach achieves up to 13.4\% improvement on MATH, 8.8\% on GPQA, and 7.0\% on GSM8K. With just two SLMs, \NAME{}  outperforms Qwen 2.5 72B on GPQA and GSM8K, and matches its performance on MATH. We further provide theoretical analyses to substantiate the advantages of our method. \homehl{Additional experiments show that the core principle of \NAME{} extends to open-ended generation tasks (e.g., HumanEval) and benefits other model classes, including frontier LLMs and domain-specific fine-tuned SLMs.} In summary, we demonstrate that SLMs can be effectively orchestrated into more accurate and efficient systems through the proposed approach. \homehl{The project page is available at \url{https://slm-mux.github.io/}.}
 
\end{abstract}


\section{Introduction}
\input{Sections/intro}
\section{Related Work}
\label{sect:related}

\input{Sections/related}


\section{Methods}
\label{sect:method}
\input{Sections/methods}
\section{Experiments}
\label{sect:experiment}

\input{Sections/experiments}
\vspace{-5pt}
\section{Discussion}
\vspace{3pt}
\label{sect:conclusion}
\input{Sections/conclusion}

\section*{Acknowledgments}
\homehl{We thank Tom Griffiths from Princeton University, whose insights and feedback were instrumental in initiating this project.}

\newpage
\bibliographystyle{iclr2026_conference}
\bibliography{iclr2026_conference}

\appendix
\input{Sections/appendix}

\end{document}

%% file: math_commands.tex
\usepackage{amsmath,amsfonts,bm}

\def\eqref#1{equation~\ref{#1}}

\def\1{\bm{1}}

\def\ra{{\textnormal{a}}}

\def\rx{{\textnormal{x}}}

\def\rva{{\mathbf{a}}}

\def\erva{{\textnormal{a}}}

\def\ervx{{\textnormal{x}}}

\def\rmA{{\mathbf{A}}}

\def\vmu{{\bm{\mu}}}
\def\vtheta{{\bm{\theta}}}
\def\va{{\bm{a}}}

\def\ve{{\bm{e}}}

\def\vx{{\bm{x}}}

\def\eva{{a}}

\def\mA{{\bm{A}}}

\def\mH{{\bm{H}}}
\def\mI{{\bm{I}}}
\def\mJ{{\bm{J}}}

\def\mX{{\bm{X}}}

\def\mSigma{{\bm{\Sigma}}}

\DeclareMathAlphabet{\mathsfit}{\encodingdefault}{\sfdefault}{m}{sl}
\SetMathAlphabet{\mathsfit}{bold}{\encodingdefault}{\sfdefault}{bx}{n}
\newcommand{\tens}[1]{\bm{\mathsfit{#1}}}
\def\tA{{\tens{A}}}

\def\tX{{\tens{X}}}

\def\gG{{\mathcal{G}}}

\def\sA{{\mathbb{A}}}
\def\sB{{\mathbb{B}}}

\def\sS{{\mathbb{S}}}

\def\emA{{A}}

\newcommand{\etens}[1]{\mathsfit{#1}}

\def\etA{{\etens{A}}}

\newcommand{\E}{\mathbb{E}}

\newcommand{\R}{\mathbb{R}}

\newcommand{\KL}{D_{\mathrm{KL}}}
\newcommand{\Var}{\mathrm{Var}}

\newcommand{\Cov}{\mathrm{Cov}}

\newcommand{\normltwo}{L^2}
\newcommand{\normlp}{L^p}

\newcommand{\parents}{Pa} 

%% file: Sections/intro.tex



Recent years have witnessed a surge of small-sized language models (SLMs) containing billions to tens of billions of parameters~\citep{wang2024comprehensivesurveysmalllanguage, javaheripi2023phi, guo2025deepseek, allal2025smollm2}. While these models may underperform state-of-the-art frontier language models, which usually contain hundreds of billions to trillions of parameters, on any given query, they offer substantially lower inference costs, are more affordable to train and fine-tune, and allow edge deployment due to their small size~\citep{belcak2025small}.  
Meanwhile, frontier models have reached trillion-parameter scales where further increases in size and training data yield diminishing returns. This mirrors a well-known challenge in computer architecture two decades ago: when enlarging single CPU cores no longer delivered proportional performance gains, computer architects turned to designing multi-core processors, where multiple smaller cores working together enabled sustained improvements.  This parallel suggests that combining multiple SLMs could offer a promising alternative to scaling ever-larger frontier models.

Recent works have explored orchestrating multiple LLMs (e.g., GPT-3.5 and GPT-4o), combining them into one system to process an input collaboratively. Representative approaches include Mixture-of-Agent~\citep{wang2024mixtureofagentsenhanceslargelanguage}, LLM-Debate~\citep{du2023improvingfactualityreasoninglanguage}, and Multi-Agent Verification~\citep{lifshitz2025multiagentverificationscalingtesttime}. These approaches share a key assumption: that models possess strong reasoning and deliberation abilities, so that interaction through natural language can reliably correct mistakes. However, when applied to SLMs, this assumption no longer holds. Our study finds that \emph{such discussion-based orchestration often fails to improve performance for SLMs}, and in some cases even reduces accuracy by over 5\%. Instead of correcting mistakes, SLMs tend to fall into groupthink during interaction, amplifying errors rather than mitigating them. The assumptions that language models can correct each other's answers behind existing orchestration methods do not hold for SLMs~\citep{Taubenfeld_2024,huang2024largelanguagemodelsselfcorrect,liu2023lostmiddlelanguagemodels,fu2025cache}.

To address this issue, we propose \textbf{\NAME{}}, a multi-model architecture for effectively orchestrating SLMs while avoiding explicit text exchanges between models. Our key insight is that \NAME{} leverages complementary abilities from different models by selecting outputs based on confidence scores without any model training.


\begin{wrapfigure}{r}{0.5\linewidth}
  \centering
  \vspace{-5pt}
  \includegraphics[width=\linewidth]{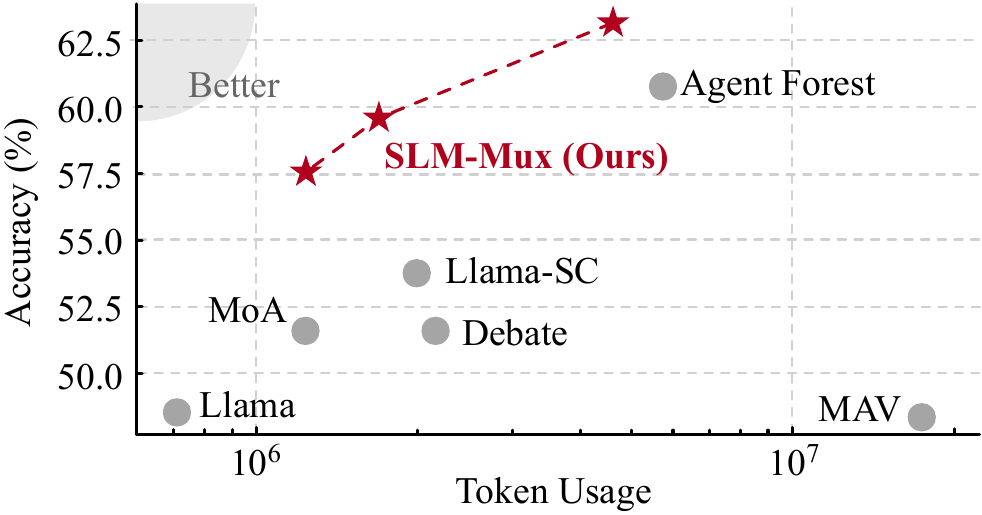}
  \vspace{-15pt}
  \caption{\small \textbf{Head-to-Head Comparison of \NAME{} with Other Methods.} \NAME{} outperforms existing methods such as Self-Consistency (SC)~\citep{wang2023selfconsistencyimproveschainthought}, Mixture-of-Agents (MoA)~\citep{wang2024mixtureofagentsenhanceslargelanguage}, LLM-Debate~\citep{du2023improvingfactualityreasoninglanguage}, Multi-Agent Verification (MAV)~\citep{lifshitz2025multiagentverificationscalingtesttime}, and Agent Forest~\citep{li2024agentsneed}. Results reported on \textsc{MATH} dataset with SLMs.}
  \label{fig:token-accuracy}
  \vspace{-20pt}
\end{wrapfigure}


After introducing \NAME{}, another question arises: which models should be orchestrated together? Not all combinations are effective -- if one model is weaker across all dimensions, it provides no benefit when paired with a stronger one. In contrast, combining models with complementary strengths (e.g., one stronger in algebra, another in geometry) allows the system to succeed where a single model would fail.

To address this, we develop a \textbf{model selection search strategy} for \NAME{}, which systematically evaluates and identifies model subsets with complementary strengths. By maximizing union accuracy while penalizing overconfident contradictions, the search procedure finds the most suitable models for a given model budget. 

In addition, we explore \textbf{compute scaling strategies} for the selected model ensembles to further enhance performance. By adjusting the number of models and samples at inference time, we further boost performance and identify practical sweet spots in the accuracy-compute tradeoff.




Our experiments demonstrate significant improvements across multiple benchmarks. By combining only two SLMs, we achieve accuracy improvements of up to 6.7\% on MATH, 5.7\% on GPQA, and 4.8\% on GSM8K, compared to the best-performing single SLMs in the system. Our method consistently outperforms existing discussion-based approaches for SLMs, with gains of up to 13.4\% on MATH, 8.8\% on GPQA, and 7.0\% on GSM8K. Most importantly,  with just two SLMs, \NAME{}  outperforms Qwen 2.5 72B on GPQA and GSM, and matches its performance on MATH.


Finally, we complement these empirical findings with theoretical and experimental analyses. Our approach shows superiority in multiple scenarios compared with previous methods (Figure \ref{fig:token-accuracy}). 

Our main contributions are as follows:
%
%
\textbf{ (i) We identify a fundamental limitation of existing orchestration methods:} Through systematic evaluation, we demonstrate that existing discussion-based methods, which show consistent improvements for frontier LLMs, actually harm performance when applied to SLMs. This counterintuitive finding challenges the assumption that orchestration methods transfer across model scales and reveals the need for SLM-specific method.
%
%
%
%
\textbf{(ii) We propose \NAME{}}, a novel multi-model architecture designed specifically for SLMs that avoids the error amplification problems of discussion-based methods. \NAME{}{} achieves consistent gains across multiple benchmarks (MATH, GPQA, GSM8K) and significantly outperforms existing discussion-based methods by large margins (up to 11.6\% on MATH).
%
%
%
\textbf{(iii) We develop principled optimization strategies} for the \NAME{}, including model selection search that identifies complementary model selections and compute scaling strategies, further boosting performance while maintaining efficiency.

%% file: Sections/related.tex
\begin{figure}[t!]
    \centering
    \includegraphics[width=\textwidth]{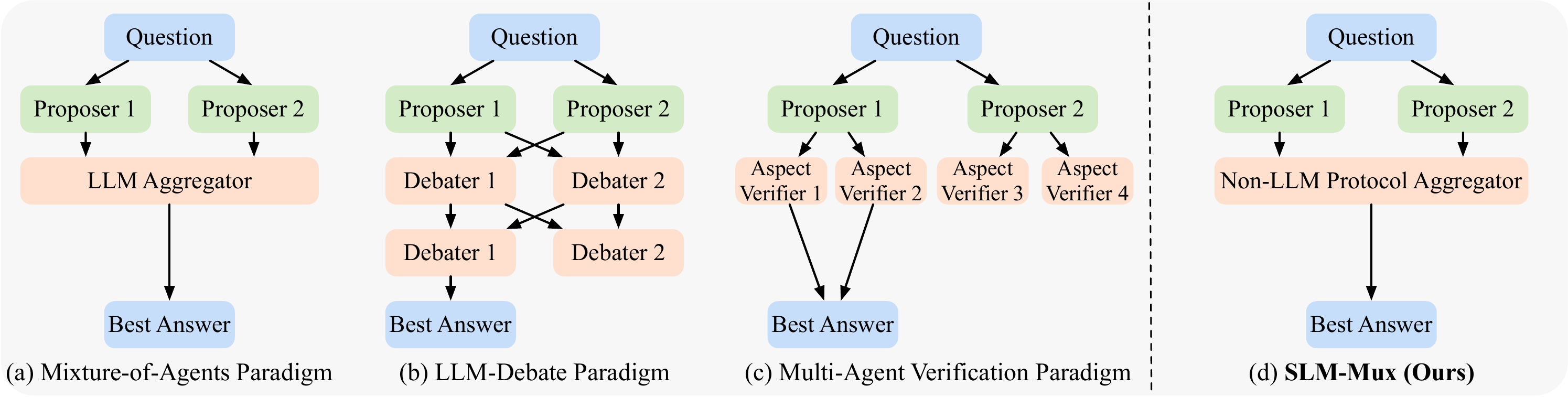}
    \vspace{-10.5pt}
    \caption{\small \textbf{Comparing \NAME{} (Ours) with Existing LLM Orchestration Methods.} (a) Mixture-of-Agents, (b) LLM-Debate, (c) Multi-Agent Verification, (d) \NAME{} (Ours).}
    
    \label{fig:comparison-methods}
\end{figure}

\paragraph{Discussion-based Orchestration Methods} We use discussion-based orchestration to refer to orchestration schemes where multiple LM instances exchange or evaluate natural-language messages~\citep{fu2025cache}—such as proposing answers, critiquing or debating, verifying from different aspects, and finally aggregating into one output. Representative approaches include Mixture-of-Agents~\citep{wang2024mixtureofagentsenhanceslargelanguage}, which uses a dedicated LLM to aggregate outputs from several models; LLM-Debate~\citep{du2023improvingfactualityreasoninglanguage}, where models critique and refine each other’s reasoning; and Multi-Agent Verification~\citep{lifshitz2025multiagentverificationscalingtesttime}, which assigns models to independently evaluate candidate solutions before selecting the final answer. These methods assume that participating models have sufficient reasoning ability to self-correct through interaction. Prior evaluations have been conducted on frontier LLMs, while their effectiveness for SLMs remains unstudied.

\looseness=-1
\paragraph{Optimization for Multi-LM Orchestration} Given these orchestration methods, some works study how to further improve their performance—e.g., how to select models to include, how to optimize prompts, or how to adapt the architecture for specific tasks~\citep{chen2023frugalgptuselargelanguage,ong2025routellmlearningroutellms,chen2024routerdcquerybasedrouterdual}. Prompt and workflow optimization methods~\citep{khattab2023dspycompilingdeclarativelanguage,opsahlong2024optimizinginstructionsdemonstrationsmultistage,saadfalcon2025archonarchitecturesearchframework, zhang2025aflowautomatingagenticworkflow} generally assume strong instruction-following ability, which makes them less effective for smaller models with limited such capabilities..

Another line of work is model selection for orchestration~\citep{chen2025optimizingmodelselectioncompound,poon2025online}. These methods often select models based on accuracy, assuming that combining models with higher standalone accuracy will yield stronger orchestrations. However, most selection criteria are not end-to-end: they evaluate models independently without directly assessing the performance of the overall orchestration. This overlooks how models interact with each other—overconfident but incorrect predictions from one model can dominate and suppress correct predictions from others, meaning that the best standalone models may not yield the best orchestration.

\paragraph{Test-time Scaling Strategies}
Test-time scaling methods improve performance by using additional computation during inference without retraining~\citep{snell2024scalingllmtesttimecompute,muennighoff2025s1simpletesttimescaling,zhang2025surveytesttimescalinglarge}. A common single-model approach is self-consistency~\citep{Trad_2025,thirukovalluru2024atomicselfconsistencybetterlong,chow2024inferenceawarefinetuningbestofnsampling}, which draws multiple samples from one model and selects the majority answer; accuracy typically improves as the number of samples increases. Agent Forest~\citep{li2024agentsneed} extends this idea to multiple models by collecting one output from each model and applying majority voting across all answers.

%% file: Sections/methods.tex



In this work, we set out to ask two critical questions: given a pool of available SLMs, how can we (i) orchestrate their outputs to achieve the best overall performance, and (ii) select an effective subset of models that maximizes accuracy?

To answer question (i), we present the \NAME{}{} (Section~\ref{sub:method-arch}), a simple yet effective orchestration method. To answer question (ii), we propose model selection search (Section~\ref{sub:method-search}) that identifies complementary subsets from dozens of available SLMs. Finally, we explore compute scaling strategies (Section~\ref{sec:scaling}) to further enhance the reasoning accuracy during inference.


\subsection{\NAME{}{} for Orchestrating Multiple Small Language Models
}
\label{sub:method-arch}

At a high level, our intuition is that we do not need to let SLMs discuss with each other. Instead, we can develop a simple rule-based method that estimates the confidence of each model’s answer and then selects the final output from the model with the highest confidence.  We term our method \textbf{\NAME{}{}}, which operates in two phases.

\textbf{Independent Generation Phase.} For a given question, we first let each SLM independently generate multiple candidate responses to the same query prompt with temperature $>0$, producing a pool of sampled answers per model.

\looseness=-1
\textbf{Confidence Estimation Phase.} We evaluate the confidence of each SLM’s outputs by measuring their consistency across their own outputs. Intuitively, a model that places higher probability mass on the correct answer will reproduce equivalent answer across samples, whereas an uncertain model will produce varied outputs. For instance, if SLM A produces three equivalent answers while model B produces three different ones, the answers from model A are more consistent and should be selected. This correlation between consistency and correctness is observed by previous papers.~\citep{wang2023selfconsistencyimproveschainthought,xie2024calibratingreasoninglanguagemodels,Taubenfeld_2025,chen2023universalselfconsistencylargelanguage}, \homehl{and we empirically revalidate this observation in Appendix~\ref{sec:acc_vs_consistency}}.

In cases where two SLMs are equally consistent but disagree, we use their validation accuracy as a tie-breaker. Prior work has shown that consistency is strongly correlated with correctness, which provides a rationale for this design.

\begin{figure}[t]
  \centering
  \includegraphics[width=\linewidth]{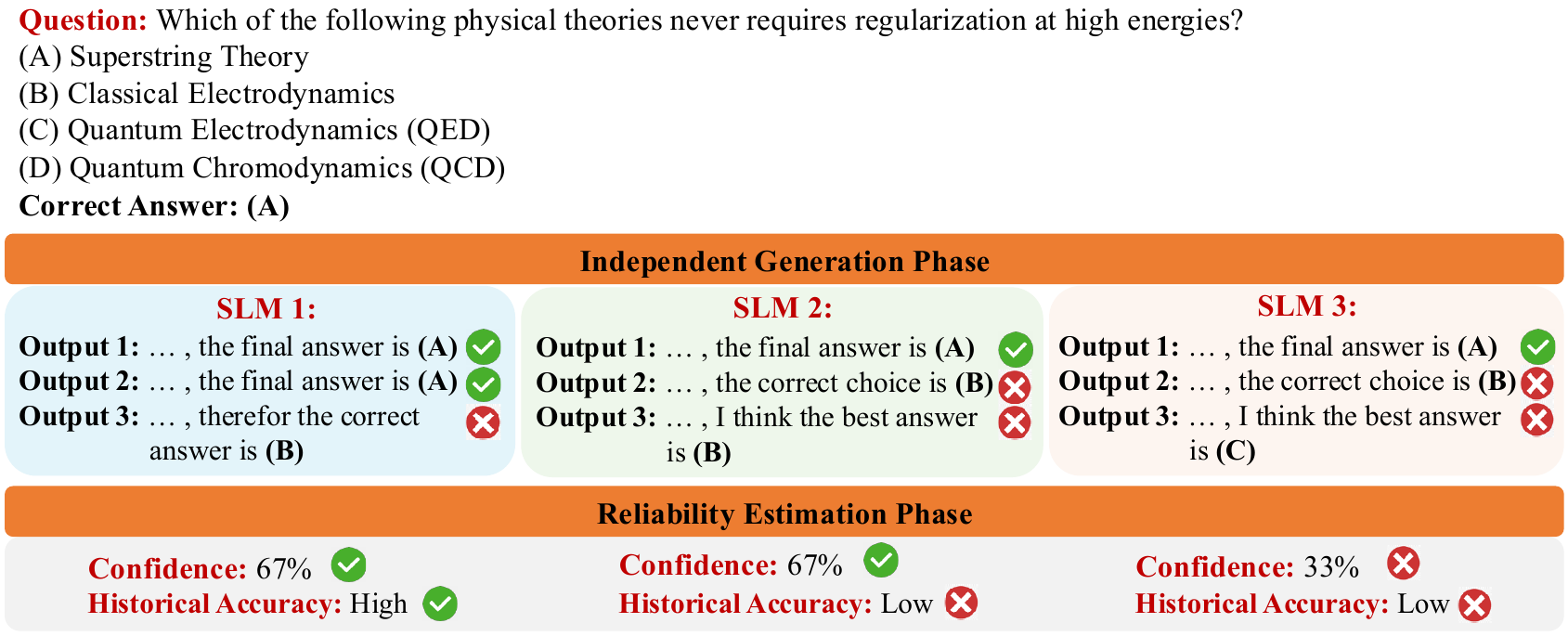}
  \vspace{-12pt}
  \caption{\small \textbf{Illustration of \NAME{}{} Workflow.}
(1) Each SLM first independently generates multiple outputs for the same question.
(2) The most frequent answer from each SLM is selected, and its frequency in the answer pool is used as the confidence score.
(3) The answers with the highest confidence score are selected.
(4) If multiple answers share the same confidence score, the tie is broken by selecting the answer from the SLM with the highest accuracy on the validation set. }
  \label{fig:SLM-Mux-method}
  \vspace{-20pt}
\end{figure}

For more details, Algorithm~\ref{alg:SLM-Mux} summarizes the workflow step by step. Figure~\ref{fig:SLM-Mux-method} provides a visual example of the workflow. The evaluation of \NAME{}{} is presented in Section~\ref{sub:vanilla}.


\begin{figure}[t]
\begin{minipage}{\linewidth}
\begin{algorithm}[H]
\caption{\NAME{}{} Working Flow}
\label{alg:SLM-Mux}
\textbf{Input}: Models $M_1,\dots,M_n$, query $x$, samples per model $k$, validation accuracies $a_1,\dots,a_n$ \\
\textbf{Output}: Final answer $\hat{y}$
\begin{algorithmic}[1]
\Statex \textit{Independent Generation: each model produces multiple candidate answers independently}
\For{$i=1,\dots,n$}
    \State Sample $k$ answers $Y_i=\{y_i^{(1)},\dots,y_i^{(k)}\}$ from $M_i$
    \State Compute $f_i(y)=\tfrac{1}{k}\sum_{j=1}^{k}\mathbf{1}\!\left(y_i^{(j)}=y\right)$
    \State Let $y_i^*=\arg\max_{y} f_i(y)$ and set $s_i = f_i(y_i^*)$
\EndFor

\Statex \textit{Confidence Estimation: measure confidence and break ties by validation accuracy}
\State $S_{\max}=\max_{i} s_i$, \quad $I^*=\{\, i \mid s_i = S_{\max} \,\}$
\If{$|I^*|=1$}
    \State $i^* \gets \text{the unique index in } I^*$
\Else
    \State $i^* \gets \arg\max_{i \in I^*} a_i$
\EndIf
\State \textbf{return} $\hat{y}=y_{i^*}^*$
\end{algorithmic}
\end{algorithm}
\end{minipage}
\vspace{-5pt}
\end{figure}


\vspace{5pt}
\subsection{Model Selection Search for SLM-MUX Optimization}
\label{sub:method-search}

%
%
%

\begin{wrapfigure}{r}{0.6\textwidth} 
\vspace{-20pt}
    \centering
    \includegraphics[width=\linewidth]{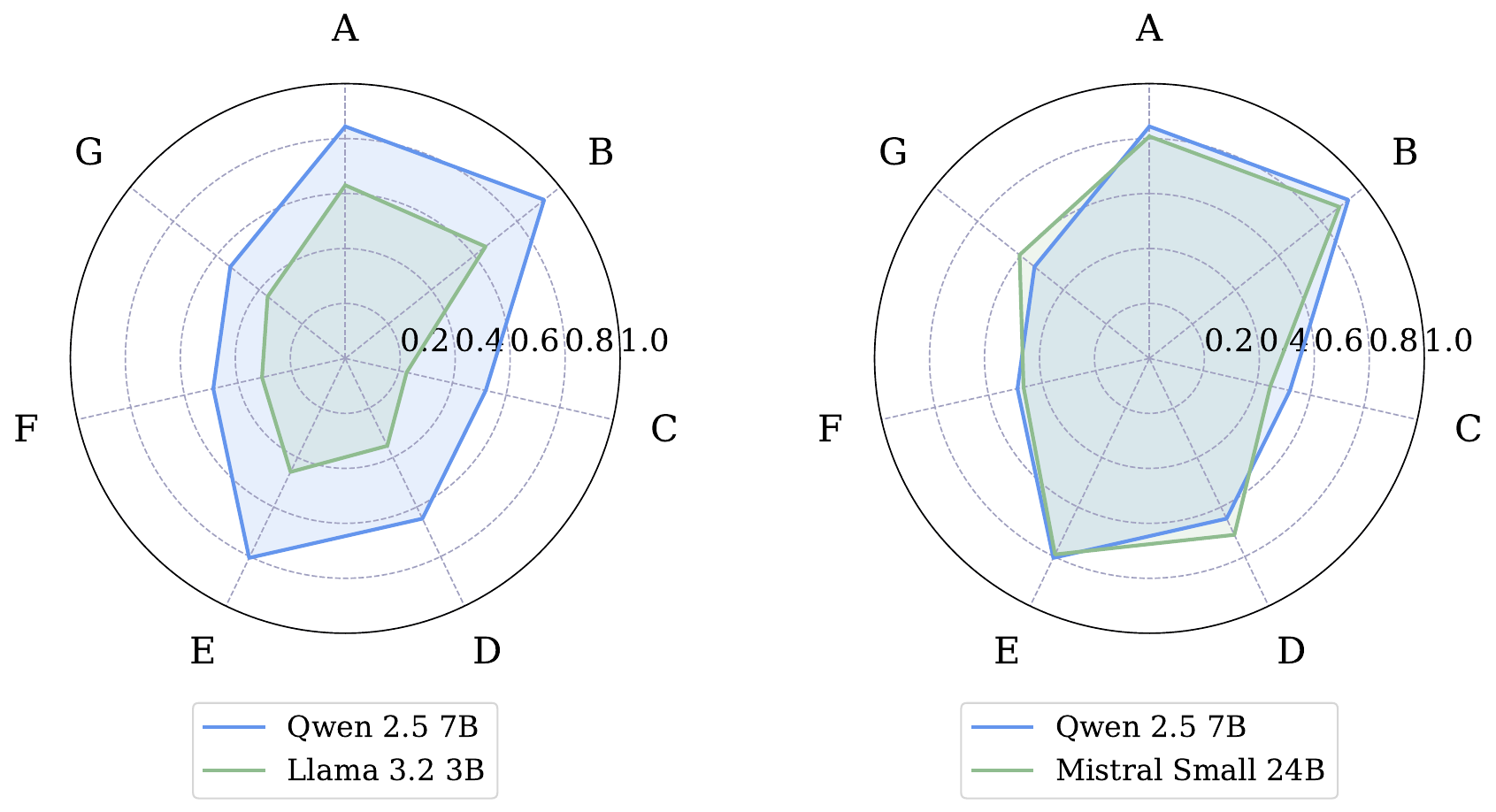}
    \vspace{-20pt}
    \caption{\textbf{Comparison of Model Choices}. Accuracy on 7 subjects for two model selection settings on MATH dataset. Subjects are denoted as: A = Prealgebra, B = Algebra, C = Intermediate Algebra, D = Number Theory, E = Counting \& Probability, F = Geometry, G = Precalculus.}

    \vspace{-15pt}
    \label{fig:motivation-search}
\end{wrapfigure}

At a high level, the idea of model selection search is to combine models with complementary skills. The goal is not simply to add more models, but to bring new capabilities as we add models. Figure~\ref{fig:motivation-search} illustrates this intuition: Qwen2.5-7B consistently outperforms Llama3.2-3B across all subjects, so combining them offers no capability beyond what Qwen2.5-7B already provides. In contrast, Mistral Small 24B and Qwen2.5-7B Mistral Small 24B and Qwen2.5-7B excels in different subjects, making their combination more effective than either model individually.

We frame model selection as a search problem on the validation set with two competing objectives. Our first objective is \textbf{Union Accuracy}, which reflects the overall accuracy 
of the system. The higher the union accuracy is, the more questions a system can potentially answer. Formally, let 
$\mathcal{M} = \{m_1, \ldots, m_K\}$ denote the set of candidate models and 
$\mathcal{D}$ the validation set. For each model $m_i \in \mathcal{M}$, we 
record the subset of validation instances it solves correctly. Given a candidate 
subset $S \subseteq \mathcal{M}$, the union accuracy is defined as
%
\begin{equation*}
\mathrm{UnionAcc}(S) =
\frac{1}{|\mathcal{D}|}
\sum_{x \in \mathcal{D}}
\mathbf{1}\!\left\{ \exists\, m \in S \;:\; m(x)\ \text{is correct} \right\}
\end{equation*}
The second objective is the \textbf{Contradiction Penalty}. It captures problematic cases where overconfident wrong answers suppress correct predictions from other models. Consider two SLMs answering the same multiple-choice question three times: the first model consistently outputs ``A'' (correct), while the second consistently outputs ``B'' (incorrect but confident). Since \NAME{}{} selects based on consistency, both models would appear equally confident, making it impossible to distinguish the correct answer from the confident but wrong one. We define this penalty as the percentage of questions where at least one model consistently gives the wrong answer while another provides the correct answer:
%
\begin{equation*}
\mathrm{Contradiction}(S) 
= \frac{1}{|\mathcal{D}|}
\sum_{x \in \mathcal{D}}
\mathbf{1}\!\Bigg\{
   \begin{array}{l}
   \exists\, m_1 \in S:\ m_1(x)\ \text{consistently wrong}, \\[4pt]
   \exists\, m_2 \in S:\ m_2(x)\ \text{correct}
   \end{array}
\Bigg\}
\end{equation*}
The final objective balances these competing factors:
\begin{equation*}
\mathcal{O}(S) \;=\; UnionAcc(S) \;-\; \lambda \cdot Contradiction(S),
\end{equation*}
Where $\lambda$ is a hyperparameter. Since the number of candidate models is not very large, 
we perform an exhaustive search. We present visualization of the two search objectives and evaluation of the searched model selection in Section~\ref{sub:search-results}.


\homehl{The rationale behind this search objective is as follows: \text{UnionAcc} represents an optimistic upper bound for \NAME{}{} performance. It assumes an ideal selection mechanism capable of identifying the correct answer whenever at least one model provides it, which is unrealistic in practice. Conversely, when $\lambda = 1$, the search objective represents a pessimistic lower bound of \NAME{}{} accuracy. This setting assumes that in cases involving confidently wrong answers, the system will invariably select the incorrect one. In practice, due to factors such as tie-breaking rules and the presence of confidently correct answers, such a worst-case scenario will not always happen. Consequently, by employing the objective $\mathcal{O}(S) = \text{Acc}(S) - \lambda \cdot \text{Contradiction}(S)$, we estimate an approximate accuracy between the theoretical upper and lower bounds of the \NAME{}{} accuracy.}

\vspace{-5pt}
\subsection{Compute Scaling Strategies}
\vspace{-3pt}
\label{sec:scaling}


Next, we empirically investigate two dimensions of test-time scaling to further enhance the performance of our \NAME{}{} with selected models.

\looseness=-1
\textbf{Adding More Participating Model Types:} As we scale the model participating model types used in the system by adding more SLMs with complementary strengths, we expect the overall accuracy to improve. For each budgeted number of models, we use the search method proposed in Section~\ref{sub:method-search} to identify the best selection from the pool. 

\textbf{Drawing More Samples per Model:} For a fixed model selection, we can increase the compute budget by scaling the number of samples drawn by each model. Since confidence is evaluated by counting the frequency of majority answers, adding more samples per model is expected to provide a more accurate confidence estimate.

These two compute scaling dimensions are evaluated in Section~\ref{sub:scaling-results}.

%% file: Sections/experiments.tex
In our experiments, we first demonstrate the fundamental limitations of existing discussion-based orchestration methods when applied to SLMs (Section~\ref{sec:comparison}). We then evaluate the proposed \NAME{} in Section~\ref{sub:vanilla}. In Section~\ref{sub:search-results}, we assess our proposed search strategy. Finally, in Section~\ref{sub:scaling-results}, we examine the compute scaling strategies.

\vspace{3pt}
\subsection{Existing Discussion-Based Orchestration Methods Harm SLM Performance}
\vspace{3pt}
\label{sec:comparison}


To understand whether orchestration methods developed for frontier LLMs are suitable for SLMs, we conduct a systematic comparison across model scales. We evaluate three prominent discussion-based methods—LLM-Debate~\citep{du2023improvingfactualityreasoninglanguage}, Mixture-of-Agents~\citep{wang2024mixtureofagentsenhanceslargelanguage}, and Multi-Agent Verification~\citep{lifshitz2025multiagentverificationscalingtesttime} —using identical experimental settings on both SLMs (Llama 3.1 8B~\citep{jiang2024mixtralexperts}, Mistral 8×7B~\citep{grattafiori2024llama3herdmodels}, Gemma 2 27B) and frontier LLMs (DeepSeek V3~\citep{deepseekai2025deepseekv3technicalreport}, Gemini 2.0 Flash~\citep{google2025gemini2flash}, GPT-4o~\citep{openai2024gpt4ocard}). Evaluation is conducted on MATH and GPQA datasets using original code and prompts.

\begin{figure}[htbp]
\vspace{-5pt}
  \centering
  \includegraphics[width=0.9\linewidth]{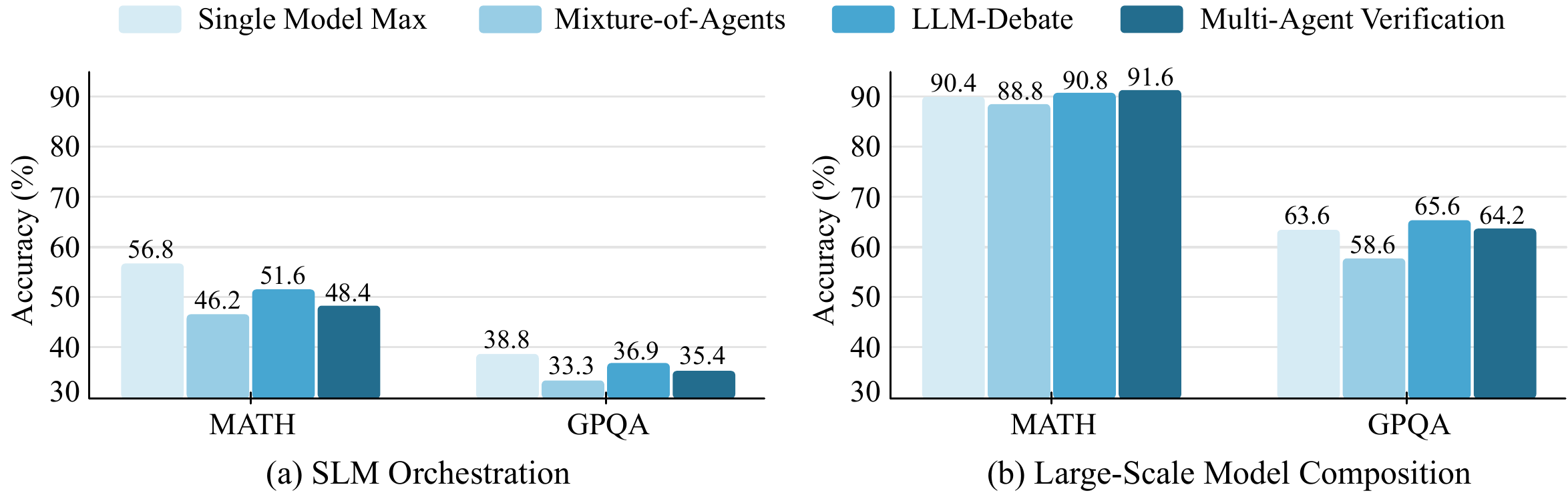}
  \vspace{-8pt}
  \caption{\small \textbf{Comparison of discussion-based orchestration when invoking SLMs and LLMs.} We compare three orchestration methods (Mixture-of-Agents, LLM-Debate, and Verification) using (a) SLMs (Llama 3.1 8B, Mistral 8$\times$7B, Gemma 2 27B) and (b) frontier LLMs (DeepSeek V3, Gemini 2.0 Flash, GPT-4o) on the \textsc{MATH} and \textsc{GPQA} datasets.  The baseline (\textit{Single-Model Max}) reflects the best performance of individual models. An orchestration is considered successful if it surpasses Single-Model Max. \homehl{All discussion-based methods are evaluated with temperature=0. The standard deviations of the accuracies are presented in Appendix~\ref{subsec:deviation}.}}\label{fig:small-large-accuracy}
  \vspace{-10pt}
\end{figure}

\paragraph{Results} As shown in Figure~\ref{fig:small-large-accuracy}, discussion-based methods generally outperform the single best-performing models in the frontier LLM group, achieving up to a 2\% increase in accuracy. However, when applied to SLMs, these discussion-based methods fail to outperform the best single model in the orchestration, and even incur accuracy drops of up to 5.5\%. This performance gap is observed across all three methods and both benchmarks. 

To understand this counterintuitive result, we analyze SLM behavior in discussion settings. We find that discussion-based methods amplify rather than correct errors in SLMs due to a key limitation: SLMs tend to exhibit groupthink, reinforcing incorrect reasoning during discussions rather than correcting mistakes. \homehl{In Appendix~\ref{sec:slm-failure}, we provide detailed analysis showing that 59.5\% of failures are attributed to groupthink, and that the performance gap persists even after extensive prompt optimization.}

\vspace{3pt}
\subsection{\NAME{} Achieves SLM Orchestration Where Existing Methods Fail}
\vspace{3pt}
\label{sub:vanilla}

\begin{wrapfigure}{r}{0.4\textwidth}
    \centering
    \vspace{-20pt}
    \includegraphics[width=.85\linewidth]{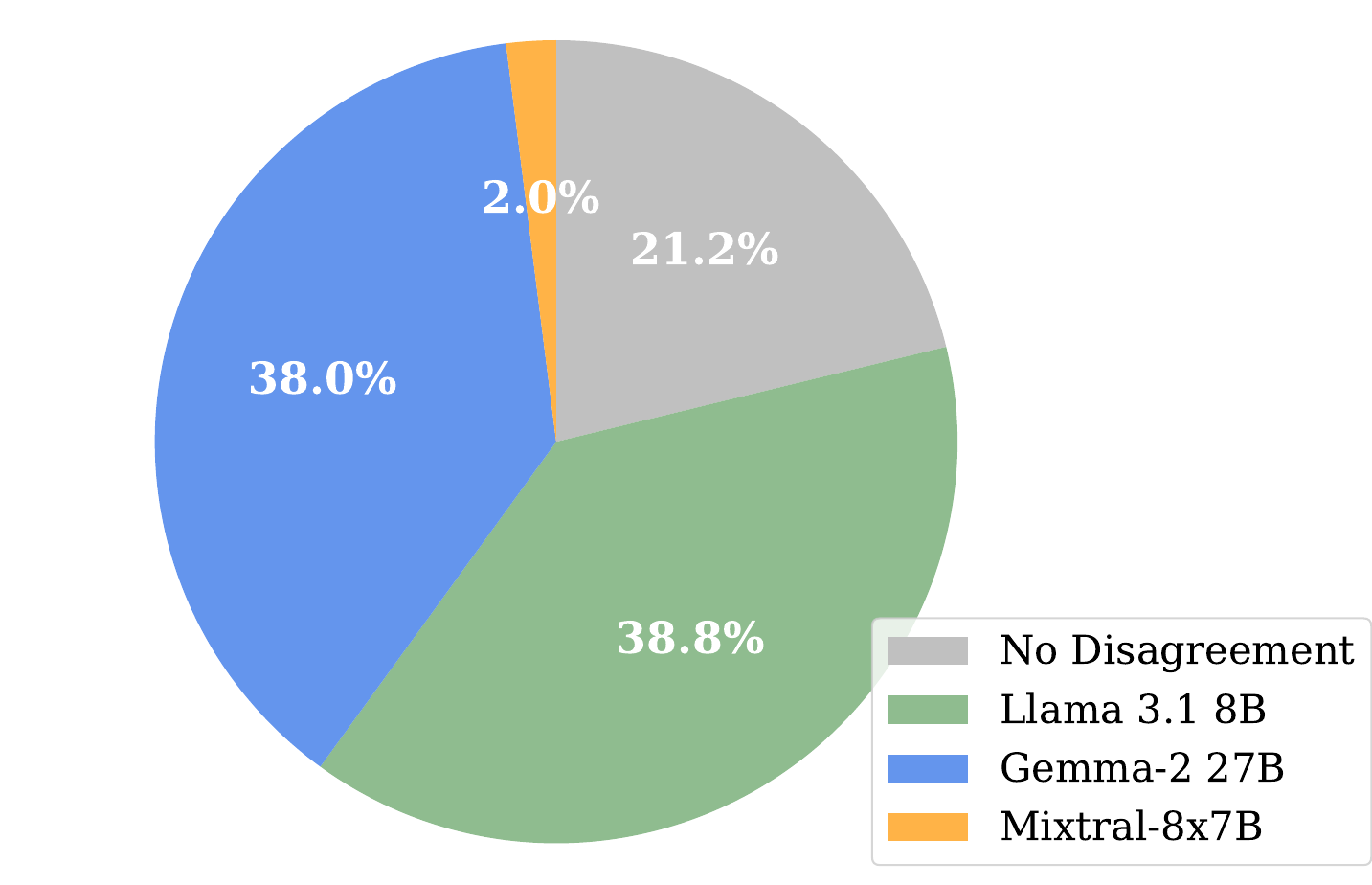}
    \caption{\textbf{Final Output Attribution}. We report the percentage of outputs contributed by each model on the MATH dataset for our \NAME{}. These results are from the same run as in Table~\ref{tab:composition-combined}.}
    \label{fig:output_percentage_math}
    \vspace{-10pt}
\end{wrapfigure}

To evaluate whether our proposed \NAME{} can successfully orchestrate SLMs, we test it against the same baselines from Section~\ref{sec:comparison}.  We use Mistral 8$\times$7B, LLaMA 3.1 8B, and Gemma 2 27B~\citep{gemmateam2024gemma2improvingopen} as base models. We implement the \NAME{} as follows. First, we generate three rounds of answers with a temperature of 0.3. Next, we compute a confidence score by counting how often the most common answer appears across these rounds. The final answer for each model is chosen as the most frequent one; in the case of a tie, we select the answer from the model with the highest validation accuracy. 

\looseness=-1
We evaluate three types of baselines. First, we measure the accuracies of individual models and report the best-performing ones. \homehl{Second, we apply self-consistency to each of the three base models independently, reporting the best-performing result as the \textit{Single-Best-SC} baseline.} Next, for comparison with existing discussion-based methods, we include LLM-Debate~\citep{du2023improvingfactualityreasoninglanguage}, Mixture-of-Agents~\citep{wang2024mixtureofagentsenhanceslargelanguage}, and Multi-Agent Verification~\citep{lifshitz2025multiagentverificationscalingtesttime}. We follow the original code and prompts described in their papers. Experiments are conducted on three benchmark datasets: MATH~\citep{hendrycks2021measuringmathematicalproblemsolving}, GPQA~\citep{rein2023gpqagraduatelevelgoogleproofqa}, and GSM8K~\citep{cobbe2021trainingverifierssolvemath}.


\looseness=-1
\paragraph{Results} Table~\ref{tab:composition-combined} summarizes the results. In our experiments, we find that for SLMs, existing orchestration methods do not consistently outperform the strongest individual base models or self-consistency approaches. In contrast, our \NAME{} generally achieves an accuracy gain. Compared with other approaches, our method yields up to a 13.4\% improvement on MATH, up to 8.8\% on GPQA, and up to 7.0\% on GSM8K.  These results demonstrate that the \NAME{} itself provides a clear advantage over alternative orchestration approaches at the architectural level. 

To better illustrate our proposed \NAME{}, we plot the output attribution for the MATH experiment (Table~\ref{tab:composition-combined}) in Figure~\ref{fig:output_percentage_math}. By selecting diverse outputs from the generation, \NAME{} leverages the complementary strengths of different SLMs.


\begin{table}[ht]
\centering
\setlength{\tabcolsep}{15pt}
\renewcommand{\arraystretch}{0.9}
\resizebox{\textwidth}{!}{
\small
\vspace{-5pt}
\begin{tabular}{lccc}
\toprule
\textbf{Method} & \textbf{MATH Acc (\%)} & \textbf{GPQA Acc (\%)} & \textbf{GSM8K Acc (\%)} \\
\midrule
Mixture-of-Agents          & 51.4 $\pm$ 2.2 & 33.3 $\pm$ 3.4 & 81.6 $\pm$ 1.7 \\
LLM-Debate                 & 51.6 $\pm$ 2.2 & 36.8 $\pm$ 3.4 & 80.8 $\pm$ 1.8 \\
Multi-Agent Verification   & 48.4 $\pm$ 2.2 & 35.3 $\pm$ 3.4 & 86.4 $\pm$ 1.5 \\
\textbf{\NAME{} (Ours)}    & \textbf{61.8 $\pm$ 1.2} & 42.1 $\pm$ 0.3 & \textbf{87.8 $\pm$ 0.6} \\
\midrule
Single-Best                & 56.8 $\pm$ 2.2 & 38.9 $\pm$ 3.5 & 84.2 $\pm$ 1.6 \\
Single-Best-SC             & 58.0 $\pm$ 2.2 & \textbf{42.4 $\pm$ 3.5} & 86.8 $\pm$ 1.5 \\
\bottomrule
\end{tabular}
}
\vspace{-5pt}
\caption{\looseness=-1 \small \textbf{Quantitative Results.} Accuracy and standard error across MATH, GPQA, and GSM8K. \homehl{``SC'' denotes self-consistency decoding (majority vote over samples from a single model), and ``Single-Best-SC'' reports the highest accuracy among the three base models when each applies self-consistency individually.}}
\label{tab:composition-combined}
\vspace{-10pt}
\end{table}

\subsection{Model Selection Search Boosts \NAME{} Performance}
\label{sub:search-results}





\begin{figure}[hbtp]
\vspace{-10pt}
    \centering
    \includegraphics[width=\textwidth]{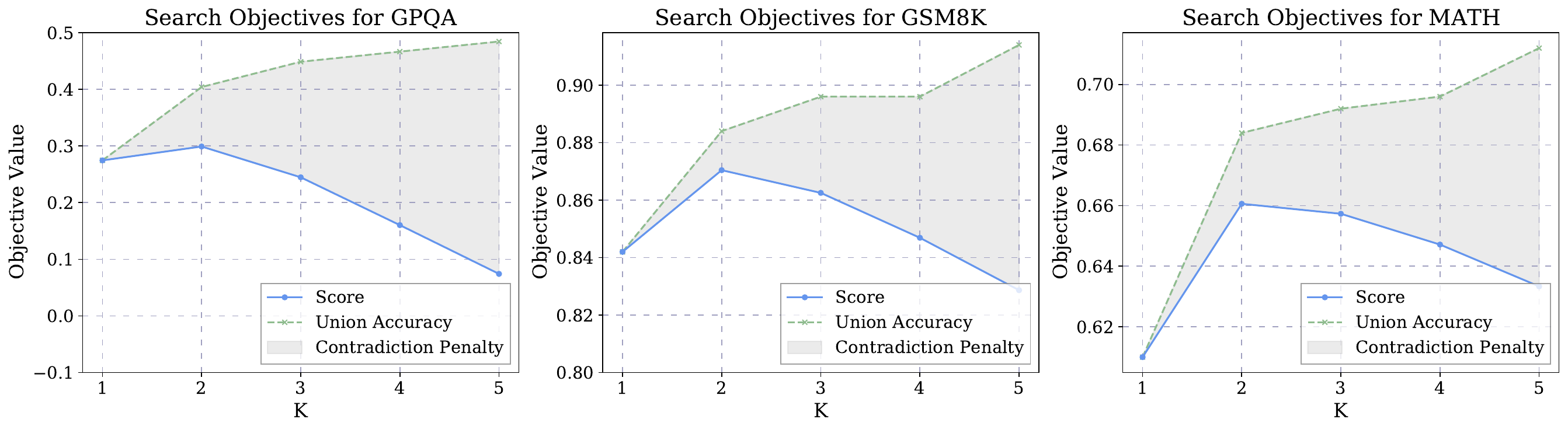}
    \vspace{-20pt}
    \caption{\textbf{Union Accuracy and Contradiction Penalty both Increases as more models are added}. We plot the search objectives as the number of models (K) increases from 2 to 5 across three benchmarks. The green line denotes the union accuracy across models, the grey area indicates the contradiction penalty, and the blue line represents the overall search objective score. \homehl{For each value of $K$, the plotted quantities are computed for the single model combination that maximizes our model selection objective defined in Section~\ref{sub:method-search}.}}
    \vspace{-5pt}
    \label{fig:search_objectives}
\end{figure}



\looseness=-1
To examine whether model selection search benefits \NAME{}, we construct a validation set of 500 questions sampled from the training splits of MATH, GPQA, and GSM8K. The candidate pool consists of five SLMs: Gemma 2 27B, Llama 3.1 8B, Mistral Small 24B~\citep{mistral_small_24b_instruct}, Mixtral 8$\times$7B, and Qwen2.5 7B~\citep{qwen2025qwen25technicalreport}. For each question, we collect three independent generations per model with temperature 0.5, repeating this process three times to obtain stable accuracy estimates.
The search procedure considers orchestrations with $K=2$ to $5$ models and is guided by an objective function mentioned in Section~\ref{sect:method}, with hyperparameter $\lambda=1$. The behavior of this objective is illustrated in Figure~\ref{fig:search_objectives}, showing the trade-off as $K$ increases. For simplicity, we select two representative two-model combinations from the search results for evaluation on the test set.

\paragraph{Results} Table~\ref{tab:composition_results_combined} summarizes the outcome of the search. The table lists the top-performing two-model combinations identified on the validation set, along with their evaluation on the held-out test set. Across benchmarks, these optimized orchestrations yield consistent improvements over the strongest individual models: accuracy increases by 4.5\% on MATH, 4.4\% on GPQA, and 4.3\% on GSM8K. This contrasts with Section~\ref{sub:vanilla}, where naive three-model combinations provide little to no benefit on GPQA. Figure~\ref{fig:search_objectives} further illustrates the underlying trade-off: while union accuracy rises with additional models, the contradiction penalty also grows, emphasizing that effective orchestration requires balancing these competing factors rather than simply enlarging the orchestration size. \homehl{In Appendix~\ref{sec:search-analysis}, we show that the \NAME{} architecture itself yields consistent gains even with randomly selected model combinations; the search procedure provides an effective and data-efficient way to further boost accuracy.}

\begin{table*}[t]
\centering
\setlength{\tabcolsep}{18pt}
\renewcommand{\arraystretch}{0.9}
\resizebox{\textwidth}{!}{
\small
\begin{tabular}{l c l c c c}
\toprule
\textbf{Benchmark} & \textbf{Group} & \textbf{Model Selection} & \begin{tabular}[c]{@{}c@{}}\textbf{Best Single}\\ \textbf{(Acc. \%)}\end{tabular} & \begin{tabular}[c]{@{}c@{}}\textbf{Composed}\\ \textbf{(Acc. \%)}\end{tabular} & \begin{tabular}[c]{@{}c@{}}\textbf{$\Delta$}\\ \textbf{(Gain)}\end{tabular} \\
\midrule
\textbf{MATH} & 1 & \makecell[l]{Mistral Small 24B \\ Qwen2.5 7B} & $75.5 \pm 1.5$ & $80.0 \pm 0.7$ & $+4.5$ \\
\cmidrule(l){2-6}
              & 2 & \makecell[l]{Qwen2.5 7B \\ Llama 3.1 8B} & $75.5 \pm 1.5$ & $77.7 \pm 0.7$ & $+2.2$ \\
\midrule
\textbf{GPQA} & 1 & \makecell[l]{Gemma 2 27B \\ Mistral Small 24B} & $45.1 \pm 2.8$ & $49.5 \pm 1.8$ & $+4.4$ \\
\cmidrule(l){2-6}
              & 2 & \makecell[l]{Llama 3.1 8B \\ Mistral Small 24B} & $45.1 \pm 2.8$ & $48.8 \pm 0.8$ & $+3.6$ \\
\midrule
\textbf{GSM8K} & 1 & \makecell[l]{Mistral Small 24B \\ Qwen2.5 7B} & $88.5 \pm 0.7$ & $92.8 \pm 0.6$ & $+4.3$ \\
\cmidrule(l){2-6}
               & 2 & \makecell[l]{Llama 3.1 8B \\ Mixtral 8$\times$7B} & $80.8 \pm 2.1$ & $85.2 \pm 0.7$ & $+4.4$ \\
\bottomrule
\end{tabular}
}
\vspace{-5pt}
\caption{\small \textbf{Model Selection Search and Evaluation Results.} We show the top two model groups identified by our search for each benchmark. For each group, we report the accuracy of the best-performing single model within the orchestration, the accuracy achieved by our \NAME{}, and the resulting performance gain.}
\label{tab:composition_results_combined}
\vspace{-3pt}
\end{table*}

\vspace{3pt}
\subsection{Compute Scaling Strategies Reveal Optimal Resource Allocation}

\label{sub:scaling-results}

\begin{figure}[t]
    \centering
    \includegraphics[width=\textwidth]{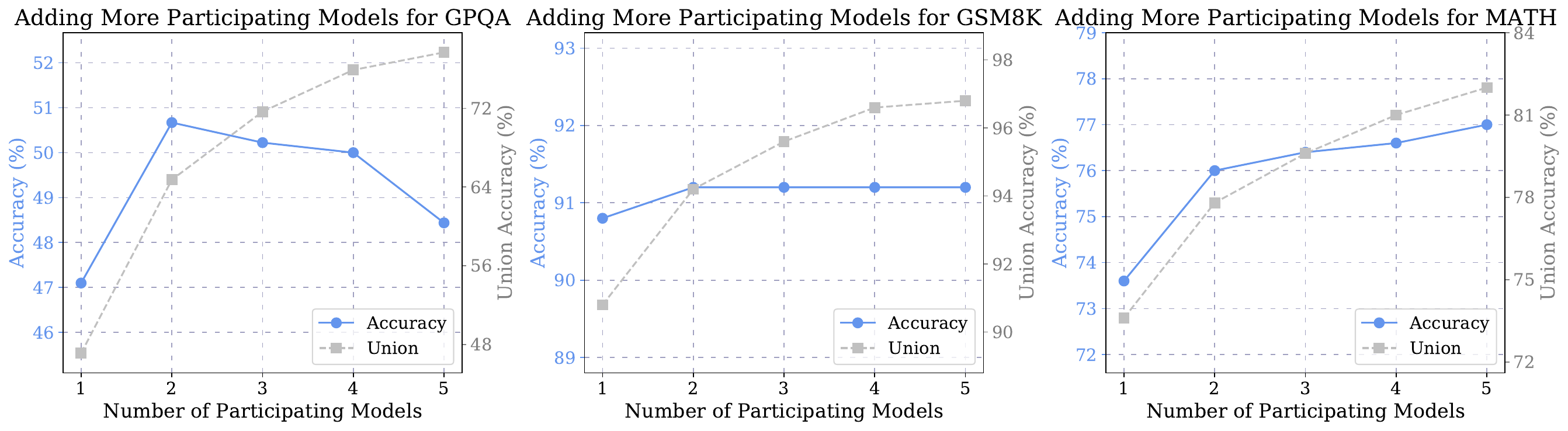}
    \vspace{-20pt}
    \caption{\small \textbf{Adding More Participating Models Affects Accuracy Differently}. We report the mean accuracy (blue line) of the optimal \NAME{}  obtained when using 2 to 5 models across three benchmarks. We also report the union accuracy (grey line), defined in Section~\ref{sub:method-search}. The blue line (Mean Accuracy) is plotted against the left-hand Y-axis. The grey line (Union Accuracy) is plotted against the right-hand Y-axis. \homehl{For each $K$, both curves correspond to the single model combination that maximizes our model selection objective (Section~\ref{sub:method-search}) under that fixed $K$.}}
    \label{fig:scaling_model_counts}
\end{figure}

To evaluate the ``Adding More Participating Model Types" dimension of compute scaling, we assess how performance changes as the number of models in the orchestration increases. For each number of models from 2 to 5, we first apply the search method from Section~\ref{sub:method-search} to identify the optimal model selection from our pool. We then evaluate \NAME{} with selected models on the validation set. Figure~\ref{fig:scaling_model_counts} plots the resulting mean accuracy (blue line, left y-axis) for each value of K. To illustrate the theoretical performance ceiling of each ensemble, we also plot the union accuracy (grey line, right y-axis), defined as the percentage of questions solved by at least one model in the group. \homehl{For each value of $K$ in Figure~\ref{fig:scaling_model_counts}, we show the single model combination that achieves the highest value of our model selection objective from Section~\ref{sub:method-search}; the search procedure is used to find the best combination under a fixed $K$, rather than to choose $K$ itself.}

\begin{table}[ht]
\vspace{-5pt}
\centering
\setlength{\tabcolsep}{20pt}
\renewcommand{\arraystretch}{0.9}
\resizebox{\textwidth}{!}{
\small
\begin{tabular}{l c c c c}
\toprule
\textbf{Benchmark} & \textbf{Samples} & \textbf{\NAME{}} & \textbf{Agent Forest} & \textbf{$\Delta$ (Gain)} \\
\midrule
\multirow{2}{*}{MATH} & 2 & $76.8 \pm 0.7$ & $72.3 \pm 1.5$ & +4.5 \\
& Best & $79.5 \pm 0.4$ & $79.2 \pm 0.4$ & +0.3 \\
\midrule
\multirow{2}{*}{GPQA} & 2 & $46.3 \pm 2.3$ & $40.4 \pm 2.3$ & +5.9 \\
& Best & $48.8 \pm 1.2$ & $47.6 \pm 1.4$ & +1.2 \\
\midrule
\multirow{2}{*}{GSM8K} & 2 & $82.1 \pm 0.7$ & $77.7 \pm 0.2$ & +4.4 \\
& Best & $86.5 \pm 0.8$ & $84.3 \pm 0.8$ & +2.2 \\
\bottomrule
\end{tabular}
}
\vspace{-5pt}
\caption{\textbf{Comparison of \NAME{} and Agent Forest.} We compare \NAME{} and Agent Forest in two settings: \textbf{(1)} with 2 samples per model (Samples=2), and \textbf{(2)} using the best accuracy found during scaling for each method (Samples=best). In the second setting, the number of samples per model may vary. }
\vspace{-5pt}
\label{tab:SLM-Mux_vs_af_comparison}
\end{table}

\begin{figure}[t]
    \centering
    \includegraphics[width=\textwidth]{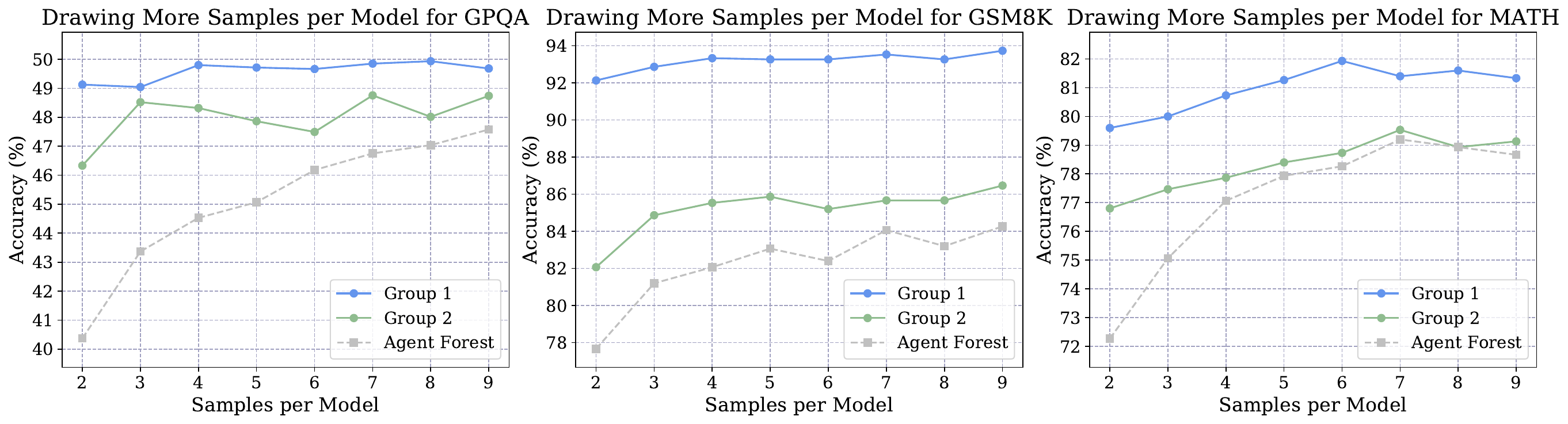}
    \vspace{-20pt}
    \caption{\small \textbf{Drawing More Samples per Model Improves Accuracy}. We report mean accuracy of \NAME{} as the number of samples per model increases from 2 to 9 across three benchmarks. Group 1 and Group 2 are from  Table~\ref{tab:composition_results_combined}. We also plot the mean accuracy of Agent Forest~\citep{li2024agentsneed} in grey line. }
    \vspace{-10pt}
    \label{fig:samples_per_model}
\end{figure}

For the ``Drawing More Samples per Model'' dimension, we reuse the two groups of models listed in Table~\ref{tab:composition_results_combined}. We vary the number of samples per model from 2 to 9 and report the mean accuracy of \NAME{} over three runs for each sample budget. The results are presented in Figure~\ref{fig:samples_per_model}, along with a baseline, Agent Forest~\citep{li2024agentsneed}, for comparison. To ensure fairness, Agent Forest is reproduced using the same models from Group 2. We report the best accuracy achieved by the \NAME{} when scaling with Samples per Model and compare it to the accuracy of the single best model in the orchestration, as shown in Table~\ref{tab:composition_results_combined}.

\paragraph{Results} The effect of ``Adding More Participating Model Types'' varies substantially across benchmarks. On GPQA, accuracy peaks when combining two models and declines thereafter. On GSM8K, accuracy quickly saturates at two models without further gains. In contrast, on MATH, accuracy continues to improve as additional models are included. Despite these differences, the union accuracy of model orchestration consistently increases with more models, emphasizing the role of output contradictions among models, as elaborated in Section~\ref{sub:method-search}. \homehl{We also validate this scaling behavior on the test set; see Appendix~\ref{app:scaling_test} for details.}

``Drawing More Samples per Model'' yields more consistent improvements across benchmarks. Moreover, under this setting, our \NAME{} systematically outperforms Agent Forest, with the largest margin observed on GPQA, where single-model accuracy is lowest. 




\begin{table}[ht]
\centering
\vspace{-10pt}
\setlength{\tabcolsep}{15pt}
\renewcommand{\arraystretch}{1.0}
\resizebox{\textwidth}{!}{
\small
\begin{tabular}{l@{\hspace{2em}}cc@{\hspace{2em}}cc@{\hspace{2em}}c}
\toprule
\multirow{2}{*}{\textbf{Benchmark}} & \multicolumn{2}{c}{\textbf{Group 1}} & \multicolumn{2}{c}{\textbf{Group 2}} & \multirow{2}{*}{\textbf{Qwen-2.5 72B Acc. \%}} \\
\cmidrule(lr){2-3} \cmidrule(lr){4-5}
& Acc. \% & $\Delta$ (Gain) & Acc. \% & $\Delta$ (Gain) &  \\
\midrule
MATH  & $81.9 \pm 0.2$ & $+6.4$ & $79.5 \pm 0.4$ & $+4.0$ & $82.3 \pm 0.5$ \\
GPQA  & $49.9 \pm 1.8$ & $+4.8$ & $48.7 \pm 1.2$ & $+3.6$ & $44.9 \pm 0.5$ \\
GSM8K & $93.7 \pm 0.2$ & $+5.2$ & $86.5 \pm 0.8$ & $+5.7$ & $90.4 \pm 0.3 $ \\
\bottomrule
\end{tabular}
}
\vspace{-5pt}
\caption{\textbf{Best Accuracy after Sample Scaling beats Larger Model.} 
Acc indicates the highest accuracy achieved through scaling. 
Groups 1 \& 2 are defined in Table~\ref{tab:composition_results_combined}. 
Gain represents the improvement over the best single-model accuracy reported in Table~\ref{tab:composition_results_combined}. 
For reference, we also include the performance of the large model Qwen-2.5 72B, showing that our composed small models can outperform it on GPQA and GSM8K.}
\label{tab:scaling_vs_qwen}
\vspace{-5pt
}
\end{table}

%% file: Sections/conclusion.tex
\vspace{-5pt}
\paragraph{\homehl{Mathematical Intuition behind SLM-MUX}} 
\homehl{Different SLMs have complementary strengths: for any given question, some models are more likely to answer correctly than others. \NAME{} exploits this by selecting the most self-consistent model's output through a simple rule-based mechanism that requires no inter-model communication.}

\homehl{The key insight is that the confidence score can identify the strongest model. We assume that for each question, there is a unique correct answer, while incorrect answers are scattered rather than clustered. Under this assumption, a model with higher accuracy $p_i$ produces the correct answer more frequently across $N$ samples, leading to a higher confidence score. Therefore, selecting the model with the highest confidence score effectively identifies the model most likely to be correct.}

\homehl{More formally, consider $K$ models where model $i$ has probability $p_i$ of being correct. Let $i^* = \arg\max_i p_i$ denote the strongest model with margin $\gamma = p_{i^*} - \max_{j \neq i^*} p_j > 0$. Under our assumption, the confidence score $s_i$ (the frequency of the most common answer over $N$ samples) concentrates around $p_i$. Applying Hoeffding's inequality and a union bound, the probability of correctly selecting the strongest model satisfies:}
\[
\Pr(\hat{i} = i^*) 
\geq 1 - 2(K-1)\exp\left(-\frac{N\gamma^2}{2}\right).
\]
\homehl{This bound shows that the probability of misidentifying the strongest model decays exponentially with sample size $N$.}

\homehl{This selection mechanism explains why \NAME{} outperforms alternatives. Unlike a single fixed model, \NAME{} performs per-question routing, effectively achieving accuracy $p_{\max}$ by always selecting the strongest available expert. Unlike pooling methods such as Agent Forest that aggregate outputs from all models, \NAME{} avoids interference from weaker models. For instance, if the strongest model has $p_1 = 0.8$ and a weaker one has $p_2 = 0.3$, pooling their outputs merely dilutes the correct answer's frequency. By isolating the strongest model and selecting its most frequent answer, \NAME{} preserves the full predictive power of the most reliable source. We provide a more detailed comparative analysis with self-consistency and Agent Forest in Appendix~\ref{sec:voting-comparison}.}

\homehl{\paragraph{Extending \NAME{} to Open-Ended Generation.}}
\homehl{Although the current implementation of \NAME{} relies on majority voting and is therefore restricted to tasks with discrete answer spaces, the underlying idea of selecting the most self-consistent model is more general. For open-ended generation, one can replace majority voting with alternative consistency estimators, such as LLM-as-a-judge scoring or embedding-based similarity measures. In Appendix~\ref{sec:humaneval}, we show a simple extension of \NAME{} to HumanEval~\citep{chen2021evaluatinglargelanguagemodels} using this idea and observe strong empirical gains.}

\homehl{\paragraph{Extending \NAME{} Beyond Generalist SLMs.}}
\homehl{The experiments above focus on general-purpose SLMs. We further evaluate whether the consistency-based selection principle extends to other settings: (1) frontier LLMs such as GPT-4o and Gemini-2.5-Flash, and (2) domain-specific fine-tuned models such as code and math specialists. In both cases, \NAME{} achieves consistent improvements over the best single model. Full experimental details are provided in Appendix~\ref{sec:beyond-slms}.}

\paragraph{Limitation and Future Work} The \NAME{} framework has two main limitations. First, its design is static and does not adapt to specific questions. For every query, it uses a fixed group of models that are pre-selected through exhaustive search -- a method that is slow and costly when there are many models to choose from. When models are tied, the framework uses their past accuracy on a validation set to decide, which is also a fixed, non-adaptive rule. Second, the way the framework measures model confidence is simple. It relies only on self-consistency -- how often a model produces the same answer. This can be a problem because a model can be very consistent while still being incorrect.

\paragraph{Conclusion} 
This work demonstrates that orchestration methods designed for frontier models paradoxically degrade the performance of SLMs by amplifying errors. To address this, we propose \NAME{}, a framework that avoids inter-model discussion, instead selecting the most reliable output based on each model's self-consistency. We further introduce a model selection search algorithm to find complementary model combinations. Experiments show our method not only substantially outperforms existing strategies but also enables an ensemble of just two SLMs to surpass the much larger Qwen-2.5 72B model on key reasoning benchmarks. In summary, our work validates that intelligently orchestrating multiple efficient models—a "multi-core" approach—is a promising alternative to scaling monolithic models on the path toward more capable AI systems.


%% file: Sections/appendix.tex
\newpage

\section*{Appendix Overview}

\homehl{The appendix is organized into five main sections.
\textbf{Section~\ref{sec:usage}} states the usage of LLMs in preparing this paper.
\textbf{Section~\ref{sec:exp-details}} provides experimental details, including visual illustrations, single-model accuracies, and standard deviation calculations.
\textbf{Section~\ref{sec:slm-failure}} analyzes why discussion-based methods fail on SLMs, presenting groupthink analysis and prompt sensitivity studies.
\textbf{Section~\ref{sec:validation}} validates the \NAME{} design through consistency-accuracy correlation analysis, comparative analysis with voting-based methods, model selection search analysis, and test-set scaling validation.
\textbf{Section~\ref{sec:generalization}} demonstrates the generalization of \NAME{} to open-ended generation, frontier LLMs, and domain-specific models.
Finally, Section~\ref{sec:licenses} provides dataset licenses.}

\section{LLM Usage Statement}
\label{sec:usage}

We used Cursor for coding. Large language models (LLMs) were employed to help polish drafts written by humans, and to assist in searching for related papers. The final choice of related work included in this paper was made entirely by the human authors after careful screening. LLMs were also used for proofreading and for providing suggestions.

\section{\homehl{Experimental Details}}
\label{sec:exp-details}

\subsection{Visual Illustrations of \NAME{}}
\label{sec:visual}

\begin{figure}[ht]
    \centering
    \includegraphics[width=0.7\linewidth]{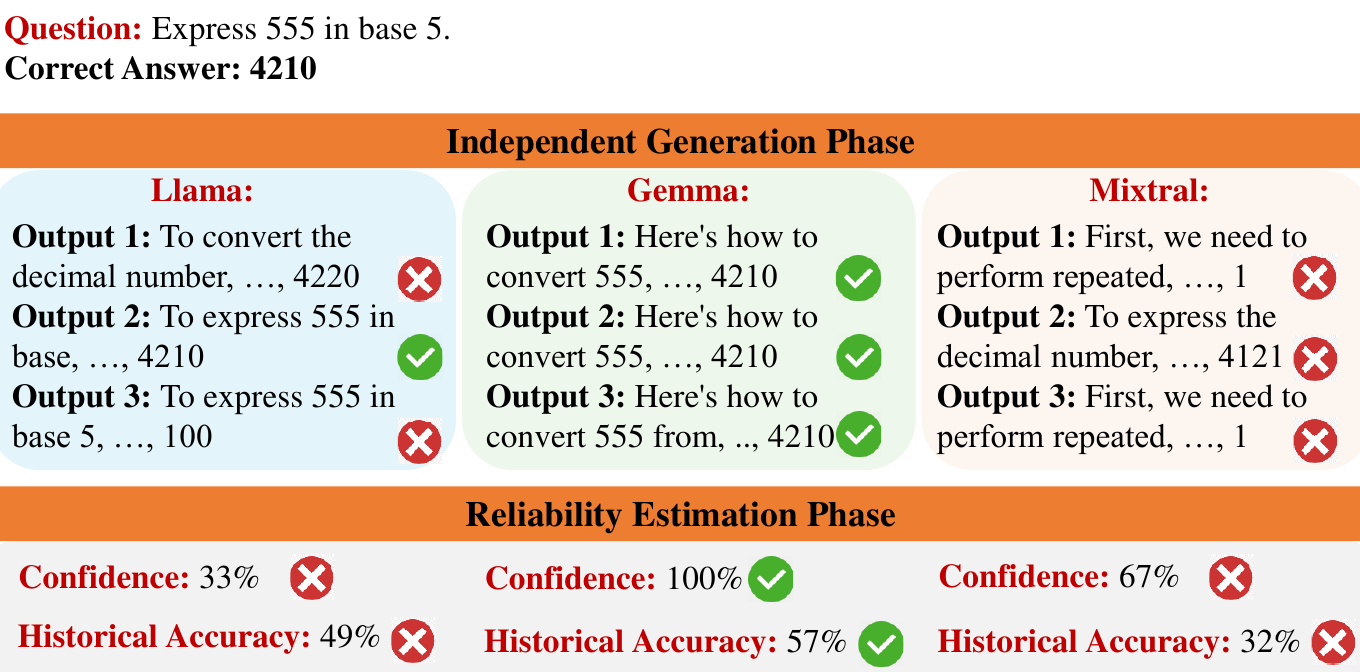}
    \caption{\textbf{An illustration of the \NAME{} method applied to the MATH dataset.} In the independent generation phase, three models are used: LLaMA-3.1-8B (denoted as Llama), Gemma-2-27B (denoted as Gemma), and Mixtral-8$\times$7B (denoted as Mixtral). Because the three models provide different answers at first, so each model is invoked two more times. Gemma obtains the highest confidence score and is therefore selected as the final output.}
    \label{fig:example_1}
\end{figure}

\begin{figure}[ht]
    \centering
    \includegraphics[width=0.7\linewidth]{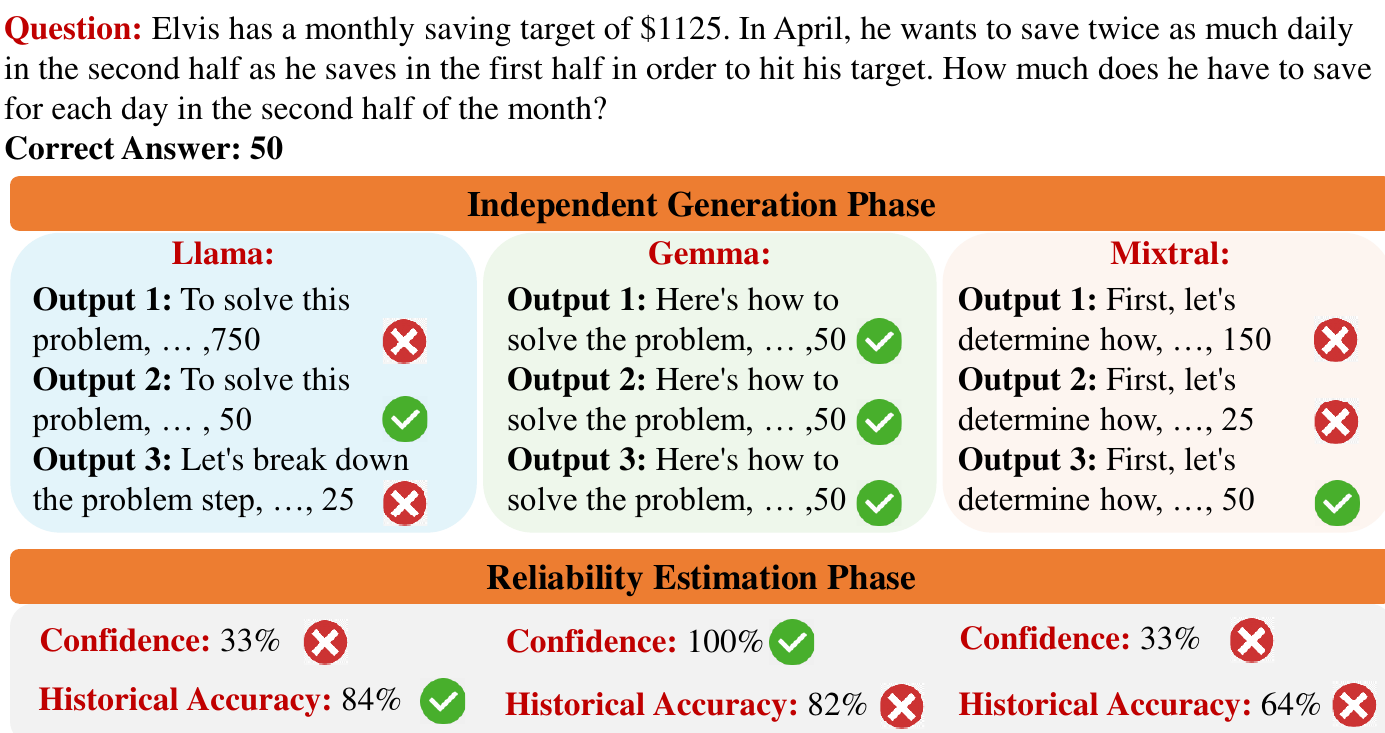}
    \caption{\textbf{An illustration of the \NAME{} method applied to the GSM8K dataset.} In the independent generation phase, different models produce different answers. However, when we invoke each model multiple times, we observe that Llama and Mixtral only yield correct answers approximately one-third of the time. In contrast, Gemma demonstrates stable performance.}
    \label{fig:example_2}
\end{figure}

\begin{figure}[ht]
    \centering
    \includegraphics[width=0.7\linewidth]{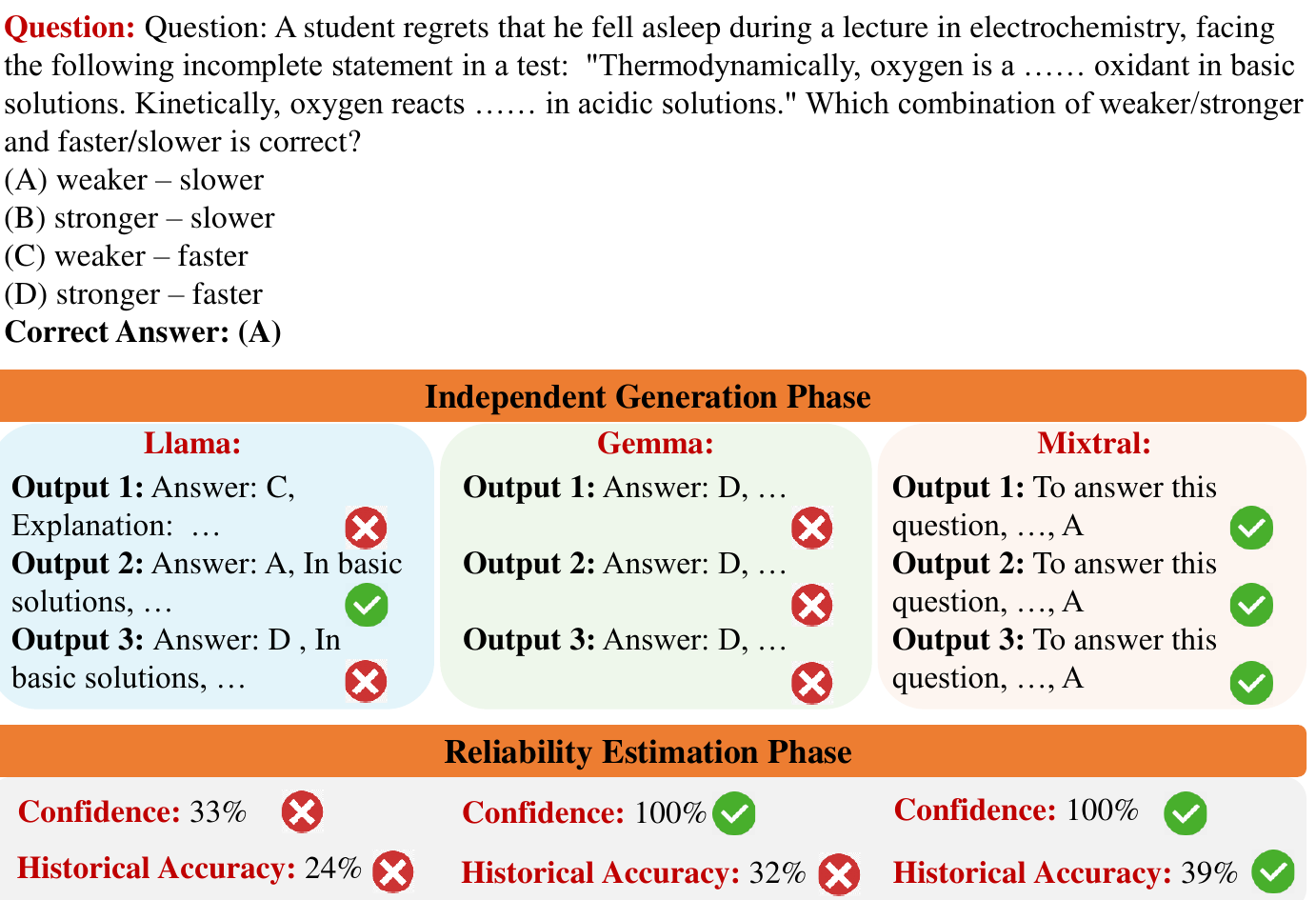}
    \caption{\textbf{An illustration of the \NAME{} method applied to the GPQA dataset.} During the independent generation phase, Gemma and Mixtral obtain the same confidence score. However, considering historical accuracy, Mixtral ranks higher. Therefore, Mixtral's answer is selected as the final output.}
    \label{fig:example_3}
\end{figure}

To more effectively illustrate the workflow of our proposed composition method, we select several representative examples from the logs. We demonstrate them in Figure~\ref{fig:example_1}, Figure~\ref{fig:example_2} and Figure~\ref{fig:example_3}.

\paragraph{\NAME{} surpasses majority voting in scenarios with initial disagreement among models.} As illustrated by Figure~\ref{fig:example_1}, during the independent generation phase, Gemma-2-27B is the sole model to provide the correct answer. Hence, majority voting applied directly would fail to select the correct author.

\subsection{Accuracy of Single LLMs}
\label{sec:single-models}

We evaluated the accuracy of single model accuracy under the condition of temperature equal to zero. The results are shown in Table~\ref{tab:base-models-combined} and Table~\ref{tab:base-models}.

\begin{table}[hbtp]
\centering
\begin{tabular}{lccc}
\toprule
\textbf{Model} & \textbf{MATH Acc (\%)} & \textbf{GPQA Acc (\%)} & \textbf{GSM Acc (\%)} \\
\midrule
Llama-3.1-8B         & 48.6 & 23.7 & 84.2 \\
Mistral-8$\times$7B  & 31.6 & 31.9 & 63.4 \\
Gemma-2-27B          & 56.8 & 38.8 & 81.6 \\
\bottomrule
\end{tabular}
\vspace{5pt}
\caption{\textbf{Small Model Base Performance.} Base model accuracy on MATH, GPQA, and GSM8K.}
\label{tab:base-models-combined}
\end{table}

\begin{table}[hbtp]
\centering
\begin{tabular}{lcccc}
\toprule
\multirow{2}{*}{\textbf{Model}} & \multicolumn{2}{c}{\textbf{MATH}} & \multicolumn{2}{c}{\textbf{GPQA}} \\
\cmidrule(lr){2-3} \cmidrule(lr){4-5}
& \textbf{Accuracy (\%)} & \textbf{Token Usage} & \textbf{Accuracy (\%)} & \textbf{Token Usage} \\
\midrule
DeepSeek V3       & 87.0  & 419,513   & 55.1  & 173,885 \\
Gemini 2.0 Flash  & 90.4  & 361,737   & 63.6  & 195,576 \\
GPT-4o            & 79.8  & 408,410   & 51.0  & 212,037 \\
\bottomrule
\end{tabular}
\vspace{5pt}
\caption{\textbf{Large Model Base Performance.} Base model performance and token usage on MATH and GPQA datasets. Accuracy is the percentage of correct answers, and token usage reflects total tokens consumed (prompt + response) over the entire dataset for each model.}
\label{tab:base-models}
\end{table}

\subsection{\homehl{Standard Deviation of the Data Points in Figure~\ref{fig:small-large-accuracy}}}
\label{subsec:deviation}

\homehl{Although all experiments are run in a deterministic setting with temperature set to zero, we can still compute the standard deviation of each datapoint by treating the outcome as a Bernoulli variable. Specifically, if there are $n_{\text{correct}}$ correct answers and $n_{\text{wrong}}$ incorrect answers, the standard deviation is}
\[
\homehl{\frac{\sqrt{\operatorname{Var}(X)}}{\sqrt{n_{\text{total}}}}
= \frac{\sqrt{p(1-p)}}{\sqrt{n_{\text{total}}}}
= \frac{\displaystyle \sqrt{\frac{n_{\text{correct}}}{n_{\text{total}}}\left(1 - \frac{n_{\text{correct}}}{n_{\text{total}}}\right)}}{\sqrt{n_{\text{total}}}},}
\]
\homehl{where $n_{\text{total}} = n_{\text{correct}} + n_{\text{wrong}}$.}

\homehl{The results are summarized in Table~\ref{tab:bernstein-std}.}

\begin{table}[hbtp]
  \centering
  \small
  \caption{\homehl{Accuracy and estimated standard deviation on MATH ($n=500$) and GPQA ($n=196$) using datapoints from Figure~\ref{fig:small-large-accuracy}.}}
  \label{tab:bernstein-std}
  \begin{tabular}{lcccc}
    \toprule
    & \multicolumn{2}{c}{\homehl{MATH ($n=500$)}} & \multicolumn{2}{c}{\homehl{GPQA ($n=196$)}} \\
    \cmidrule(lr){2-3} \cmidrule(lr){4-5}
    \homehl{Method} & \homehl{SLM orchestration} & \homehl{LLM composition} & \homehl{SLM orchestration} & \homehl{LLM composition} \\
    \midrule
    \homehl{Single Model Max}
      & \homehl{$56.8 \pm 2.22$} & \homehl{$90.4 \pm 1.32$}
      & \homehl{$38.8 \pm 3.48$} & \homehl{$63.6 \pm 3.44$} \\
    \homehl{Mixture-of-Agents}
      & \homehl{$46.2 \pm 2.23$} & \homehl{$88.8 \pm 1.41$}
      & \homehl{$33.3 \pm 3.37$} & \homehl{$58.6 \pm 3.52$} \\
    \homehl{LLM-Debate}
      & \homehl{$51.6 \pm 2.23$} & \homehl{$90.8 \pm 1.29$}
      & \homehl{$36.9 \pm 3.45$} & \homehl{$65.6 \pm 3.39$} \\
    \homehl{Multi-Agent Verification}
      & \homehl{$48.4 \pm 2.23$} & \homehl{$91.6 \pm 1.24$}
      & \homehl{$35.4 \pm 3.42$} & \homehl{$64.2 \pm 3.42$} \\
    \bottomrule
  \end{tabular}
\end{table}

\section{Why Discussion-Based Methods Fail on SLMs}
\label{sec:slm-failure}

\subsection{Groupthink Analysis}
\label{subsec:groupthink}

We analyze the experiment logs of LLM-Debate using small language models (SLMs) in Section~\ref{sec:comparison}. Among 500 debate problems, 242 resulted in failure (48.4\%). For each of the 242 failed debates, we first used an analyzer LLM to produce a process-focused failure analysis. We then used a separate labeling LLM to classify whether each failed debate was due to groupthink.

The labeling results are shown in Table~\ref{tab:label}:

\begin{table}[t]
\centering
\begin{tabular}{@{}l r l@{}}
\toprule
Metric & Count & Rate \\
\midrule
Total Debates Analyzed & 500 & 100\% of total \\
Failed Debates (System Error) & 242 & 48.4\% of total \\
\midrule
\multicolumn{3}{@{}l}{\textit{Breakdown of Failed Debates:}} \\
\quad Attributed to Groupthink & 144 & 59.5\% of failures \\
\quad Attributed to Other Causes & 79 & 32.6\% of failures \\
\quad Classification Unsuccessful & 19 & 7.9\% of failures \\
\bottomrule
\end{tabular}
\caption{\textbf{Failure Cause Attribution} This table shows the cause attribution for LLM-Debate when involving SLMs.}
\label{tab:label}
\end{table}

These results reinforce our claim that groupthink is a major failure mode in SLM-based LLM-debate.

We provide the exact prompts used by (i) the analyzer LLM to generate the 242 failure analyses (Figure~\ref{lst:prompt-analyzer}) and (ii) the groupthink labeler LLM to classify groupthink (Figure~\ref{lst:prompt-groupthink-system}). Placeholders such as \texttt{\{problem\}} indicate runtime substitutions by our code.

\lstdefinestyle{promptstyle}{basicstyle=\ttfamily,breaklines=true,backgroundcolor=\color{gray!10}}

\begin{figure}[htbp]
  \begin{minipage}{\linewidth}
  \begin{lstlisting}[style=promptstyle]
As an expert in analyzing multi-agent AI systems, your task is to analyze why an 'LLM Debate' process failed to find the correct answer. Your focus should be on the *debate dynamics and process*, not just the mathematical details. The goal is to understand the failure of the debate methodology itself.

**Ground Truth:**
- **Problem Statement:** {problem}
- **Correct Answer:** {ref_answer}

**Debate Information:**
- **Final Incorrect Answer from System:** {system_answer}

**Analysis of Round 1:**
- **Model `{model_name}` proposed:**
  - Answer: `{extracted_answer}`
  - Reasoning:
```
{full_text}
```

... (repeats per round and per model)

**Your Analysis Task:**
Based on the debate history, provide a "Debate Failure Analysis". Do not focus on simple calculation mistakes. Instead, analyze the interaction between the models and the structure of the debate. Pinpoint the core reasons the *debate process* failed. Consider these questions:

1.  **Error Propagation vs. Correction:** How did initial errors influence later rounds? Were there moments where a correct idea was introduced but ignored or overruled? Why did the debate fail to self-correct?
2.  **Groupthink and Influence Dynamics:** Did the models converge on a flawed consensus? Did one or more influential but incorrect models lead the group astray? Was there evidence of independent reasoning that was shut down?
3.  **Argumentation Quality:** Did the models provide convincing but ultimately flawed arguments? Did they effectively challenge each other's reasoning, or was the debate superficial?
4.  **Critical Failure Point in the Debate:** Identify the single most critical turn or moment in the debate that sealed its failure. What happened, and why was it so impactful?
5.  **Improving the Debate:** What is the single most important change to the debate protocol or dynamics that could have prevented this failure? (e.g., different communication rules, promoting dissident opinions, etc.)

Provide a concise, expert analysis focusing on the *process* failure.
\end{lstlisting}
  \end{minipage}
  \caption{Prompt Template for Failure Analysis.}
  \label{lst:prompt-analyzer}
\end{figure}

\begin{figure}[htbp]
  \begin{minipage}{\linewidth}
  \begin{lstlisting}[style=promptstyle]
You are an expert analyst of multi-agent LLM debates. Your goal is to determine whether the failure primarily involved groupthink/conformity dynamics. Groupthink indicators include: early flawed consensus, explicit capitulation to a majority, social proofing, adopting peers' answers without critique, abandoning independent reasoning to match others, or reinforcing an incorrect majority despite available dissent. Not-groupthink includes failures due to independent arithmetic/logic errors, argument complexity/veneer effects without convergence, or chaotic divergence with no consensus influence. Return STRICT JSON only, with keys: groupthink (bool), confidence (float 0-1), reasons (string), cues (array of strings).
\end{lstlisting}
  \end{minipage}
  \caption{Prompt for Groupthink Classification.}
  \label{lst:prompt-groupthink-system}
\end{figure}

\subsection{\homehl{Prompt Sensitivity Analysis}}
\label{subsec:prompt-sensitivity}

\homehl{A natural concern is whether the performance gap between discussion-based methods and \NAME{} is due to suboptimal prompt selection rather than inherent limitations. To address this, we conduct a prompt sensitivity analysis on the Mixture-of-Agents (MoA) baseline using the MATH benchmark.}

\homehl{We use Gemini-2.5-Flash to generate 10 diverse aggregator prompts for MoA. Table~\ref{tab:prompt-sensitivity} summarizes the results. Our baseline prompt (46.2\%) outperforms the average of the tuned prompts (41.8\%) and surpasses 7 out of 10 generated prompts. Even the best tuned prompt (48.4\%) remains substantially below the best single model (56.8\%) and \NAME{} (61.8\%).}

\begin{table}[htbp]
    \centering
    \caption{\homehl{Prompt sensitivity analysis for MoA on MATH. The baseline prompt used in our experiments is already near-optimal.}}
    \label{tab:prompt-sensitivity}
    \begin{tabular}{lc}
        \toprule
        \homehl{Setting} & \homehl{Accuracy} \\
        \midrule
        \homehl{\NAME{} (Ours)} & \homehl{\textbf{61.8\%}} \\
        \homehl{Best Single Model} & \homehl{56.8\%} \\
        \midrule
        \homehl{Tuned Prompt (Best)} & \homehl{48.4\%} \\
        \homehl{Baseline Prompt (Used in Paper)} & \homehl{46.2\%} \\
        \homehl{Tuned Prompt (Average)} & \homehl{41.8\%} \\
        \homehl{Tuned Prompt (Worst)} & \homehl{25.8\%} \\
        \bottomrule
    \end{tabular}
\end{table}

\homehl{To further stress-test this result, we conducted iterative prompt optimization directly on the test set for 6 rounds, selecting the best-performing prompt at each iteration. Even under this extremely favorable setting for MoA, the accuracy peaked at 55.6\%, still below the best single model (56.8\%) and far below \NAME{} (61.8\%). This confirms that the performance gap reflects inherent limitations of discussion-based aggregation when applied to SLMs, rather than an artifact of prompt selection.}

\section{\homehl{Validation of \NAME{} Design}}
\label{sec:validation}

\subsection{\homehl{Consistency vs Accuracy Correlation}}
\label{sec:acc_vs_consistency}

\homehl{We empirically study how per-question self-consistency correlates with accuracy on four datasets: GSM8K, MATH, GPQA, and \textsc{HumanEval}. For each model–dataset pair, we compute a self-consistency score for every question (as defined in the main text) and group questions into three bins according to this score: \emph{Low} $[0.0, 0.5)$, \emph{Medium} $[0.5, 0.8)$, and \emph{High} $[0.8, 1.0]$. We then measure the empirical accuracy (fraction of correct answers) within each bin.}

\begin{figure}[htbp]
  \centering
  \includegraphics[width=0.8\linewidth]{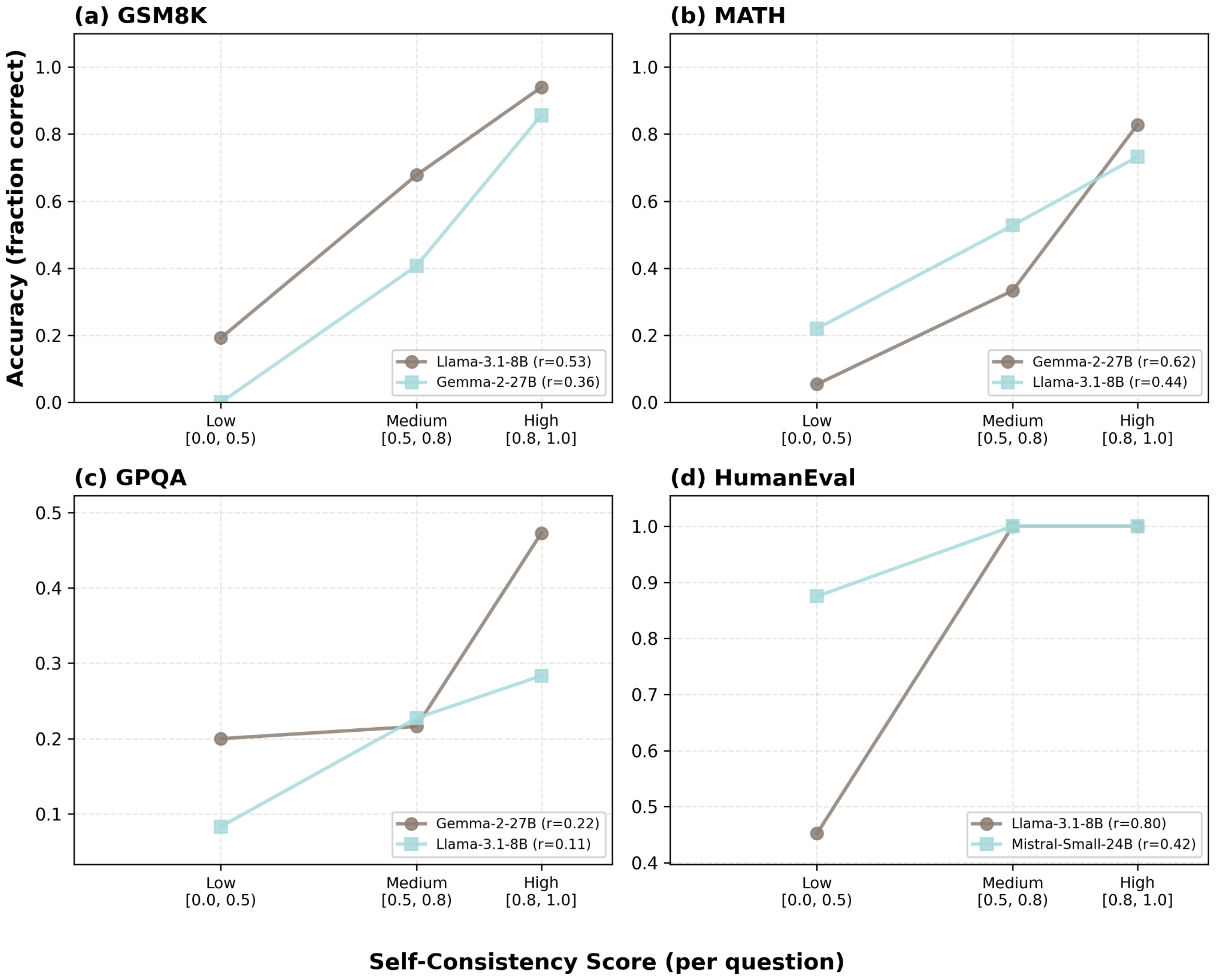}
  \caption{Accuracy as a function of per-question self-consistency score on GSM8K, MATH, GPQA, and \textsc{HumanEval}. Questions are grouped into three bins by self-consistency: \emph{Low} $[0.0,0.5)$, \emph{Medium} $[0.5,0.8)$, and \emph{High} $[0.8,1.0]$. Each line corresponds to a different SLM; legends report the Pearson correlation $r$ between self-consistency and correctness.}
  \label{fig:consistency-accuracy}
\end{figure}

\homehl{Figure~\ref{fig:consistency-accuracy} reports the resulting accuracies for two representative SLMs on each dataset. Across GSM8K, MATH, and \textsc{HumanEval} we observe a strong positive relationship between self-consistency and accuracy: questions in the high-consistency bin are substantially more likely to be answered correctly than those in the low-consistency bin. GPQA exhibits a weaker but still positive correlation. Overall, these results provide empirical support for the link between self-consistency and correctness assumed in our method.}

\subsection{\homehl{Comparative Analysis with Voting-Based Methods}}
\label{sec:voting-comparison}

\homehl{Since \NAME{} also involves voting on model outputs, we examine its differences from standard self-consistency and Agent Forest to better explain the source of our improvements.}

\homehl{To explain our stronger performance, we note a limitation of self-consistency methods. Suppose a model has probability $p$ of answering a question correctly. When self-consistency samples $N$ responses, the probability of obtaining the correct answer after majority voting follows a binomial distribution:}
\begin{equation}
\homehl{A(N, p) = \Pr\left( X \geq \left\lceil \tfrac{N}{2} \right\rceil \right) = \sum_{k=\lceil N/2 \rceil}^{N} \binom{N}{k} p^{k}(1-p)^{N-k}, \quad X \sim \text{Binomial}(N,p)}
\end{equation}
\homehl{We observe that $A(N,p)$ exceeds $p$ only when $p > 0.5$, meaning self-consistency is effective only in this regime. When $p < 0.5$, self-consistency can actually lower overall accuracy.}

\homehl{For any dataset, we can conceptually divide examples into three types of questions. \textbf{Type 1}: $p = 100\%$, the model always answers correctly. \textbf{Type 2}: $p > 50\%$, the model is more likely than not to be correct. \textbf{Type 3}: $p < 50\%$, the model is more likely to be wrong. The overall effect of self-consistency is then the improvement from Type 2, offset by the degradation from Type 3. Improvement occurs only when the dataset contains a sufficiently large proportion of Type 2 questions.}

\homehl{For \NAME{}, we select the output from the most confident model, so the accuracy can be approximated as $A(N, p_{\max})$, where $p_{\max}$ is the highest probability among the participating models. By routing to the model with the highest $p_{\max}$ on each question, we effectively enlarge the proportion of Type 2 questions, leading to higher overall accuracy.}

\homehl{For the Agent Forest approach, answers are drawn evenly from all models, so its accuracy can be approximated as $A(N, \bar{p})$, where $\bar{p}$ is the average probability across models. This generally results in lower accuracy than \NAME{}, as weaker models dilute the signal from stronger ones.}

\subsection{\homehl{Model Selection Search Analysis}}
\label{sec:search-analysis}

\subsubsection{\homehl{Search-Set Size Stability}}
\label{sec:search-set-size}

\homehl{We analyze how the size of the search set (number of problems used in the search phase) affects the resulting model ranking. For each of the three benchmarks MATH, GSM8K, and GPQA, we treat the full benchmark (approximately 500 problems) as the search pool and first run our search procedure on the full set to obtain a ranking of all candidate models. We then record the models occupying Rank 1, Rank 2, and Rank 3 under this full-set ranking.}

\homehl{Next, we subsample the search set to smaller sizes and re-run the search. Specifically, for each dataset we evaluate search-set sizes of $100$, $200$, $300$, $400$, and the full set. At each size, we recompute the ranking over all models and track the ranks of the three models that were Rank 1–3 under the full-set setting.}

\begin{figure}[htbp]
    \centering
    \includegraphics[width=\linewidth]{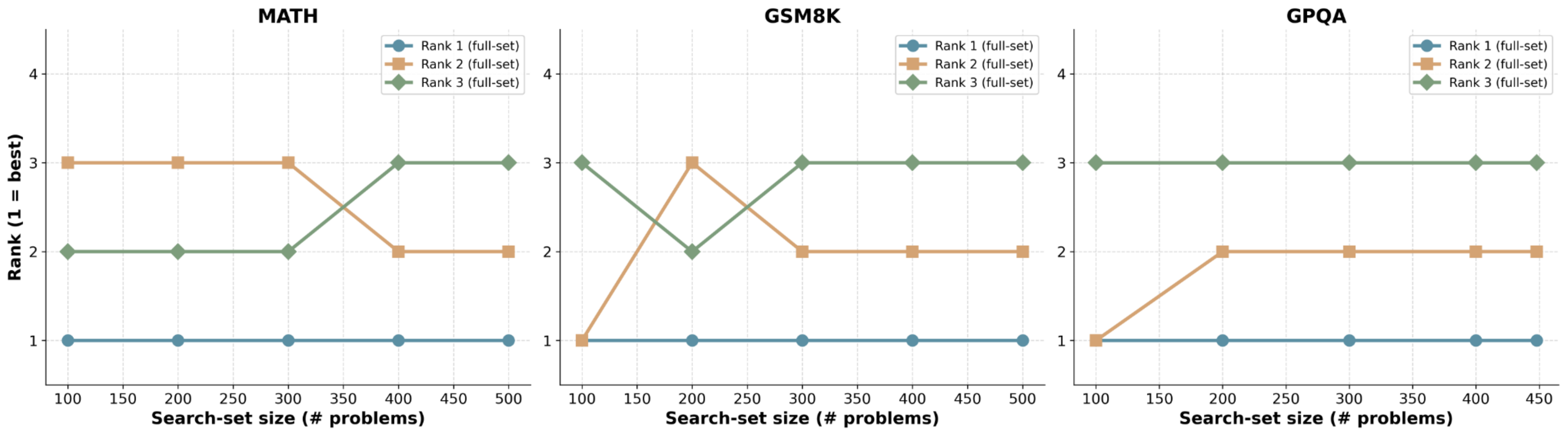}
    \caption{\homehl{Stability of model rankings with respect to search-set size on MATH, GSM8K, and GPQA. For each dataset, we first determine the top three models using the full search set and then track their ranks as the search-set size is reduced.}}
    \label{fig:search-size-stability}
\end{figure}

\homehl{Figure~\ref{fig:search-size-stability} shows the results. Each curve corresponds to one of the Rank 1/2/3 models under the full-set ranking, and the $y$-axis reports its rank when the search-set size is changed. Across all three datasets, these models consistently remain within the top three positions even when the search set is reduced to as few as 100 problems, with only minor swaps in their relative order on MATH and GSM8K. This suggests that a search set of a few hundred problems is sufficient to stably identify the top-performing models.}

\subsubsection{\homehl{Random Model Selection Baseline}}
\label{sec:random-selection}

\homehl{A natural question is whether the performance gains of \NAME{} stem from the model selection search or from the orchestration architecture itself. To isolate the contribution of the architecture, we conduct an experiment where model combinations are selected randomly rather than through our search procedure.}

\homehl{For each dataset, we randomly sample model combinations of size $K = 2, 3, 4$ from our pool of five SLMs and apply \NAME{}. We compare the resulting accuracy against the best single model within each random pool. Table~\ref{tab:random-selection} reports the average performance across all random combinations.}

\begin{table}[htbp]
    \centering
    \caption{\homehl{Performance of \NAME{} with randomly selected model combinations. Even without optimized model selection, \NAME{} consistently outperforms the best single model in the pool.}}
    \label{tab:random-selection}
    \begin{tabular}{llccc}
        \toprule
        \homehl{Dataset} & \homehl{$K$} & \homehl{\NAME{} (Random)} & \homehl{Best Single Model} & \homehl{$\Delta$ (Gain)} \\
        \midrule
        \homehl{MATH-500} & \homehl{2} & \homehl{67.2\%} & \homehl{66.1\%} & \homehl{+1.1} \\
        & \homehl{3} & \homehl{71.5\%} & \homehl{69.8\%} & \homehl{+1.7} \\
        & \homehl{4} & \homehl{73.7\%} & \homehl{71.8\%} & \homehl{+1.9} \\
        \midrule
        \homehl{GSM8K} & \homehl{2} & \homehl{84.1\%} & \homehl{81.7\%} & \homehl{+2.4} \\
        & \homehl{3} & \homehl{87.3\%} & \homehl{83.6\%} & \homehl{+3.7} \\
        & \homehl{4} & \homehl{88.8\%} & \homehl{84.0\%} & \homehl{+4.8} \\
        \midrule
        \homehl{GPQA} & \homehl{2} & \homehl{43.0\%} & \homehl{41.3\%} & \homehl{+1.7} \\
        & \homehl{3} & \homehl{44.5\%} & \homehl{44.4\%} & \homehl{+0.1} \\
        & \homehl{4} & \homehl{46.3\%} & \homehl{46.0\%} & \homehl{+0.3} \\
        \bottomrule
    \end{tabular}
\end{table}

\homehl{As shown in Table~\ref{tab:random-selection}, \NAME{} consistently outperforms the best single model even when the model combination is selected randomly. On MATH-500 and GSM8K, the gains are substantial (up to +4.8 on GSM8K with $K=4$). On GPQA, where consistency is a weaker signal for correctness, the gains are smaller but still positive.}

\homehl{To further quantify how often \NAME{} improves over single models, we compute the \textbf{Effective Combination Rate}: the percentage of all possible $K$-model subsets where \NAME{} outperforms the best single model in that subset. Table~\ref{tab:effective-rate} reports the results.}

\begin{table}[htbp]
    \centering
    \caption{\homehl{Effective Combination Rate: percentage of model combinations where \NAME{} outperforms the best single model in the subset.}}
    \label{tab:effective-rate}
    \begin{tabular}{lccc}
        \toprule
        \homehl{Dataset} & \homehl{$K=2$} & \homehl{$K=3$} & \homehl{$K=4$} \\
        \midrule
        \homehl{MATH-500} & \homehl{70\%} & \homehl{100\%} & \homehl{100\%} \\
        \homehl{GSM8K} & \homehl{100\%} & \homehl{100\%} & \homehl{100\%} \\
        \homehl{GPQA} & \homehl{60\%} & \homehl{80\%} & \homehl{100\%} \\
        \bottomrule
    \end{tabular}
\end{table}

\homehl{On GSM8K, 100\% of combinations are effective across all values of $K$. On MATH-500 and GPQA, the effective rate increases with $K$, reaching 100\% at $K=4$. Even at $K=2$, the majority of combinations (60--100\%) are effective. These results demonstrate that the space of ``workable'' model combinations is dense, and one does not need to search extensively to find an effective subset. The model selection search provides additional gains by identifying the optimal combination, but the architecture is robust even without it.}

\subsection{\homehl{Test-Set Validation of Scaling Behavior}}
\label{app:scaling_test}

\homehl{In Section~\ref{sec:scaling}, we evaluated the ``Adding More Participating Model Types'' scaling dimension on the validation set. Here we report the corresponding results on the test set to verify that the observed trends generalize.}

\begin{figure}[htbp]
    \centering
    \includegraphics[width=\linewidth]{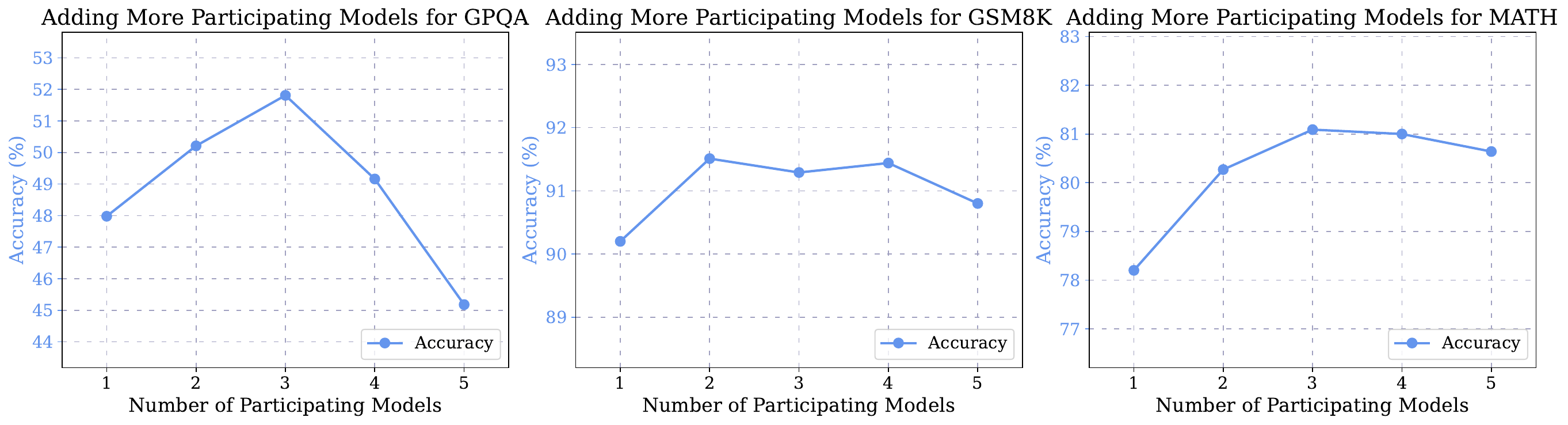}
    \caption{\homehl{Test-set accuracy as a function of the number of participating models on GPQA, GSM8K, and MATH. The trends mirror those observed on the validation set: GPQA peaks at $K=3$, GSM8K saturates quickly, and MATH shows continued improvement up to $K=3$.}}
    \label{fig:scaling_test}
\end{figure}

\homehl{Table~\ref{tab:scaling_test} summarizes the test-set accuracy as the number of participating models $K$ increases from 1 to 5. For each value of $K$, we use the model combination selected by the search procedure described in Section~\ref{sub:method-search}. Figure~\ref{fig:scaling_test} visualizes these results.}

\begin{table}[htbp]
    \centering
    \caption{\homehl{Test-set accuracy (\%) as the number of participating models $K$ increases. For $K=1$, the best single model is reported.}}
    \label{tab:scaling_test}
    \begin{tabular}{lccc}
        \toprule
        \homehl{$K$} & \homehl{GPQA} & \homehl{GSM8K} & \homehl{MATH} \\
        \midrule
        \homehl{1 (best single)} & \homehl{47.98} & \homehl{90.20} & \homehl{78.20} \\
        \homehl{2} & \homehl{50.21} & \homehl{91.51} & \homehl{80.27} \\
        \homehl{3} & \homehl{51.81} & \homehl{91.29} & \homehl{81.09} \\
        \homehl{4} & \homehl{49.16} & \homehl{91.44} & \homehl{81.00} \\
        \homehl{5} & \homehl{45.18} & \homehl{90.80} & \homehl{80.64} \\
        \bottomrule
    \end{tabular}
\end{table}

\section{\homehl{Generalization of \NAME{}}}
\label{sec:generalization}
\label{sec:beyond-slms}

\subsection{\homehl{Open-Ended Generation (HumanEval)}}
\label{sec:humaneval}

\homehl{In the main text, \NAME{} is instantiated on tasks with discrete answer spaces, where self-consistency can be measured via majority voting over sampled outputs. To check whether the same principle extends to open-ended generation, we apply \NAME{} to the \textsc{HumanEval} code-generation benchmark.}

\paragraph{\homehl{Consistency estimator for open-ended generation.}}
\homehl{On \textsc{HumanEval}, exact-string majority voting is not appropriate, so we replace it with a semantic consistency estimator. For each model and each problem, we sample $N=5$ code generations with temperature $0.3$. We then encode the 5 generations using the pretrained embedding model \texttt{Salesforce/codet5p-110m-embedding} and compute pairwise cosine similarities, yielding a $5 \times 5$ similarity matrix. From this matrix, we identify the most coherent cluster of generations (of size $x$) and use $x/5$ as the model's self-consistency score for that problem, analogous to the confidence score derived from majority voting in the discrete-answer setting.}

\homehl{We also experimented with an LLM-as-a-judge–based consistency estimator (using Qwen2.5-7B-Instruct as the judge) and found that the embedding-based estimator exhibits a stronger correlation with ground-truth correctness (Pass@1). All results below therefore use the embedding-based consistency score.}

\paragraph{\homehl{Results.}}
\homehl{Table~\ref{tab:humaneval-single} reports the Pass@1 of each individual SLM on \textsc{HumanEval}. Table~\ref{tab:humaneval-mux} reports the Pass@1 of \NAME{} when combining two models at a time; for each problem, \NAME{} selects the solution from the model with the larger embedding-based consistency score.}

\begin{table}[t]
    \centering
    \caption{\homehl{Pass@1 on \textsc{HumanEval} for individual SLMs.}}
    \label{tab:humaneval-single}
    \begin{tabular}{lc}
        \toprule
        \homehl{Model} & \homehl{Pass@1} \\
        \midrule
        \homehl{Llama-3.1-8B-Instruct}   & \homehl{0.178} \\
        \homehl{Qwen2.5-7B-Instruct}     & \homehl{0.485} \\
        \homehl{Mistral-Small-24B}       & \homehl{0.870} \\
        \homehl{Qwen2.5-Coder-7B}        & \homehl{0.893} \\
        \bottomrule
    \end{tabular}
\end{table}

\begin{table}[t]
    \centering
    \caption{\homehl{\NAME{} on \textsc{HumanEval} (Pass@1). Each row corresponds to a pair of SLMs; \NAME{} selects the output from the model with higher embedding-based consistency.}}
    \label{tab:humaneval-mux}
    \begin{tabular}{lcc}
        \toprule
        \homehl{Setup} & \homehl{Models combined} & \homehl{\NAME{} (Pass@1)} \\
        \midrule
        \homehl{Exp 1} & \homehl{Llama-3.1-8B-Instruct + Qwen2.5-7B-Instruct} & \homehl{\textbf{0.506}} \\
        \homehl{Exp 2} & \homehl{Mistral-Small-24B + Qwen2.5-Coder-7B}        & \homehl{\textbf{0.939}} \\
        \bottomrule
    \end{tabular}
\end{table}

\subsection{\homehl{Frontier LLMs}}
\label{subsec:frontier-llms}

\homehl{We evaluate whether \NAME{} can exploit complementary strengths between state-of-the-art frontier models. We pair GPT-4o with Gemini-2.5-Flash and apply \NAME{} on MATH-500, GPQA, and GSM8K-500. For each problem, we sample $N=5$ responses per model at temperature 0.3 and apply self-consistency routing: for each problem, we perform majority voting within each model's samples, then route to the model showing higher agreement.}

\homehl{Table~\ref{tab:frontier-llm} summarizes the results. As a reference, ``Perfect Routing'' indicates the theoretical upper bound achievable if the system always selects the correct model when at least one succeeds.}

\begin{table}[htbp]
    \centering
    \caption{\homehl{\NAME{} performance when applied to frontier LLMs (GPT-4o and Gemini-2.5-Flash).}}
    \label{tab:frontier-llm}
    \begin{tabular}{lcccc}
        \toprule
        \homehl{Benchmark} & \homehl{GPT-4o} & \homehl{Gemini-2.5-Flash} & \homehl{\NAME{}} & \homehl{Perfect Routing} \\
        \midrule
        \homehl{MATH-500} & \homehl{73.0\%} & \homehl{92.1\%} & \homehl{92.8\%} & \homehl{94.2\%} \\
        \homehl{GPQA} & \homehl{50.7\%} & \homehl{51.1\%} & \homehl{\textbf{60.1\%}} & \homehl{73.7\%} \\
        \homehl{GSM8K-500} & \homehl{89.1\%} & \homehl{85.7\%} & \homehl{89.4\%} & \homehl{91.4\%} \\
        \bottomrule
    \end{tabular}
\end{table}

\homehl{The results reveal two distinct regimes. On GPQA, the two models exhibit complementary error patterns, and \NAME{} achieves 60.1\% accuracy, surpassing the best single model by nearly 10 percentage points. This demonstrates that \NAME{} effectively exploits complementary strengths even at the frontier scale. On MATH and GSM8K, the Perfect Routing bounds (94.2\% and 91.4\%) are only marginally higher than the single-model baselines, indicating high overlap in the models' correct predictions. The limited gains on these benchmarks reflect this ceiling rather than a limitation of the routing mechanism.}

\subsection{\homehl{Domain-Specific Fine-Tuned Models}}
\label{subsec:domain-specific}

\homehl{Domain-specific fine-tuned models are widely deployed in practice. We evaluate whether \NAME{} can effectively orchestrate such specialized models by testing on two domains: code generation and mathematical reasoning.}

\homehl{For code generation, we pair Qwen2.5-Coder-7B (a code-specialized model) with Mistral-Small-24B (a general-purpose model) on HumanEval. We use the same embedding-based consistency estimator described in Section~\ref{sec:humaneval}. For mathematical reasoning, we pair DeepSeek-Math-7B-RL (a math-specialized model) with Llama-3.1-8B-Instruct on MATH-500, using standard majority voting for consistency estimation. Both experiments sample $N=5$ responses per model at temperature 0.3.}

\begin{table}[t]
    \centering
    \caption{\homehl{\NAME{} performance when orchestrating domain-specific fine-tuned models.}}
    \label{tab:domain-specific}
    \begin{tabular}{llccc}
        \toprule
        \homehl{Domain} & \homehl{Benchmark} & \homehl{Best Single Model} & \homehl{\NAME{}} & \homehl{$\Delta$ (Gain)} \\
        \midrule
        \homehl{Code Generation} & \homehl{HumanEval} & \homehl{89.3\%} & \homehl{\textbf{93.9\%}} & \homehl{+4.6} \\
        \homehl{Math Reasoning} & \homehl{MATH-500} & \homehl{58.8\%} & \homehl{\textbf{62.2\%}} & \homehl{+3.4} \\
        \bottomrule
    \end{tabular}
\end{table}

\homehl{As shown in Table~\ref{tab:domain-specific}, \NAME{} achieves consistent improvements in both domains. On HumanEval, the orchestrated system reaches 93.9\% Pass@1, outperforming the code specialist (89.3\%) by 4.6 percentage points. On MATH-500, combining the math specialist with a general-purpose model yields 62.2\% accuracy, a 3.4pp improvement. These results demonstrate that \NAME{} generalizes effectively to domain-specific fine-tuned models, successfully capturing complementary strengths between specialists and generalists.}

\section{Licenses for Datasets}
\label{sec:licenses}
The MATH dataset is licensed under the MIT License.\\
The GPQA dataset is licensed under the Creative Commons Attribution 4.0 International (CC BY 4.0) License.\\
The GSM8K dataset is licensed under the MIT License.